\documentclass{article}
\usepackage{graphicx} 
\graphicspath{ {./Results/} }
\usepackage{amsmath}
\usepackage{titlesec}
\usepackage{biblatex} 
\usepackage{float}
\usepackage{authblk}

\titleformat{\paragraph}
{\normalfont\normalsize\bfseries}{\theparagraph}{1em}{}
\titlespacing*{\paragraph}
{0pt}{3.25ex plus 1ex minus .2ex}{1.5ex plus .2ex}
\title{Neuronal Fluctuations: Learning Rates vs Participating Neurons}
\author[1]{Darsh Pareek}
\author[2]{Umesh Kumar}
\author[3]{Ruthu Rao*}
\author[4]{Ravi Janjam}
\affil[1]{Undergraduate Researcher, Numerikal Labs}
\affil[2]{Undergraduate Researcher, Numerikal Labs}
\affil[3]{Graduate Researcher, Numerikal Labs, *Graduate student at Purdue University}
\affil[4]{Principal Investigator / Chief of R\&D, Numerikal Labs}
\begin{document}
\maketitle
\begin{abstract}
    Deep Neural Networks (DNNs) rely on inherent fluctuations in their internal parameters—weights and biases—to effectively navigate the complex optimization landscape and achieve robust performance. While these fluctuations are recognized as crucial for escaping local minima and improving generalization, their precise relationship with fundamental hyperparameters remains underexplored. A significant knowledge gap exists concerning how the learning rate, a critical parameter governing the training process, directly influences the dynamics of these neural fluctuations. This study systematically investigates the impact of varying learning rates on the magnitude and character of weight and bias fluctuations within a neural network. We trained a model using distinct learning rates (0.01, 0.001, and 0.0001) and analyzed the corresponding parameter fluctuations in conjunction with the network's final accuracy. Our findings aim to establish a clear link between the learning rate's value, the resulting fluctuation patterns, and overall model performance. By doing so, we provide deeper insights into the optimization process, shedding light on how the learning rate mediates the crucial exploration-exploitation trade-off during training. This work contributes to a more nuanced understanding of hyperparameter tuning and the underlying mechanics of deep learning.
\end{abstract}
\newpage
\section{Introduction}
Deep neural networks have revolutionized various fields, achieving state-of-the-art performance in complex tasks such as image recognition, natural language processing, and medical diagnosis. A cornerstone of this success lies in the ability of these networks to learn intricate patterns from data through the adjustment of their internal parameters, namely the weights and biases of individual neurons. It is well-established that the process of learning is not a monotonic descent but rather involves inherent fluctuations in these parameters. These fluctuations, often viewed as a consequence of the stochastic nature of optimization algorithms like Stochastic Gradient Descent (SGD), are not merely noise but are believed to play a crucial role in enabling networks to escape local minima and explore a broader parameter space, thereby facilitating generalization and improved performance.

Despite the acknowledged importance of these parameter fluctuations, a significant gap remains in our understanding of their relationship with key hyperparameters. Specifically, the interplay between the learning rate, a fundamental parameter that dictates the step size of each weight update, and the characteristics of these neural fluctuations has not been systematically investigated. The learning rate is a critical determinant of a model's convergence and final accuracy, and it is plausible that its value profoundly influences the magnitude, frequency, and overall dynamics of weight and bias fluctuations. A low learning rate might lead to smaller, more stable fluctuations, while a high learning rate could induce more volatile, yet potentially beneficial, parameter changes.

Caffeine blocks receptors and increases neuron excitability, making you feel more alert. Dopamine changes learning speed in response to reward signals. AI is similar to adaptive learning rates and temperature scaling in softmax. The model adapts how quickly it learns or how ``confident'' it is based on the situation. Chemicals are used in biology to tune excitability, while AI uses learning-rate schedules or scaling to tune sensitivity.

Nicotine presents a different example of tuning in biology influenced by fluctuation. 
Upon binding to nicotinic acetylcholine receptors, neurons become further excitable 
\cite{Pidoplichko2004}, leading to increased attention temporarily 
\cite{Ji2023}. However, if the stimulation continues, the receptors begin to desensitize, 
and neuronal responses diminish — a normal mechanism to prevent overstimulation 
\cite{QuickLester2002}. This give and take, increasing intensity then reducing it, 
is a hallmark of nicotine’s effect on neural signaling, where the same stimulus 
can shift from enhancement to suppression depending on exposure frequency 
\cite{Mansvelder2002,Zhang2004}. 

This biological feedback process resembles the behavior of learning systems. 
They begin with noise scheduling or dropout decay, allowing randomness to encourage 
exploration initially, then gradually stabilize to consolidate what has been learned 
\cite{Burton2018}. In both the brain and artificial networks, this dynamic adjustment 
maintains flexibility while preventing instability or overfitting, achieving a 
balanced form of adaptive control \cite{Haimerl2023}.

This study aims to bridge this knowledge gap by systematically investigating the impact of different learning rates on the fluctuations of weights and biases within a neural network. Our central thesis is that by analyzing these fluctuations in conjunction with the network's accuracy, we can gain deeper insights into the optimization process and the mechanisms by which learning rates influence model performance. Through this research, we seek to understand how the learning rate shapes the exploration-exploitation trade-off during training, providing a more nuanced perspective on hyperparameter tuning and network optimization. The remainder of this paper is structured as follows: We will first describe the dataset and neural network model used in our experiments. We will then detail the range of learning rates explored and the training procedure. A key component of our methodology involves the saving and analysis of fluctuation data throughout the training process. Finally, we will present and discuss our results, linking the observed patterns of neural fluctuations to the final accuracy achieved by the network.

\subsection{Learning rate hyperparameter and it's impact on neurons demonstrated}
The learning rate is a critical hyperparameter that dictates the magnitude of weight adjustments during model training. The selection of an optimal value is essential for effective convergence. Given that the ideal value is problem-dependent, the learning rate is typically tuned empirically, often by evaluating performance on a validation set partitioned from the training data. This study investigates this process at the neuronal level by examining three distinct learning rates: 0.01, 0.001, and 0.0001. The objective is to analyze how variations in this hyperparameter affect the fluctuations of internal neuronal parameters and to determine their ultimate influence on the final output of the network.
\section{Method}
\subsection{Overview}
Our research explores the intricate dynamics of a neural network's learning process. To do this, we designed a controlled environment where we could precisely monitor how different training parameters impact the model's internal state. Our methodology, which uses a custom autoencoder, can be broken down into three main phases: dataset and model architecture, training the model under specific conditions, and meticulously capturing and analyzing the resulting data.
\subsection{Datasets and Model Architecture}
To ensure our experiments were consistent and reproducible, we created our own environment from the ground up.
\subsubsection{Generating a Controlled Dataset}
Instead of using a pre-existing dataset, we used the NumPy \cite{Numpy} library to synthesize our own \cite{SynData}. This gave us complete control over our input data, allowing us to precisely define the characteristics of the shapes our model would learn. Our dataset consists of eight distinct geometric shapes—triangle, square, pentagon, hexagon, heptagon, octagon, circle, and spiral—with 500 samples for each. Each shape is represented as a series of (x,y) coordinates, a simple format that makes it easy for our model to process.
\subsubsection{A Custom Neural Network for Fine-Grained Analysis}
We built our model using the PyTorch \cite{Pytorch} deep learning framework, which provided the flexibility we needed to create a custom neural network architecture. This custom design was crucial because it allowed us to gain fine-grained control over every internal parameter, a key requirement for our detailed analysis.
The heart of our approach is a feed-forward autoencoder \cite{Autoencoders}. This type of model is designed to compress input data and then reconstruct it, essentially learning a compact representation of the data. Our autoencoder is composed of two main parts
\paragraph{Encoder}
The encoder is responsible for compressing the 2-dimensional input data into a single-dimensional latent space representation. It consists of a series of three fully connected (linear) layers, with ReLU \cite{ReLU} activation functions applied after the first two layers. The layers are structured as follows \\ \\ Layer 1: A linear layer that takes a 2-dimensional input and outputs a 64-dimensional tensor. This is followed by a Rectified Linear Unit (ReLU) activation function.
    $$
    \text{Output}_1=ReLU(Linear(\text{Input}_{2\xrightarrow{}64}))
    $$
    Layer 2: A linear layer that takes the 64-dimensional output from the previous layer and outputs a 32-dimensional tensor. This is also followed by a ReLU activation function.
    $$
    \text{Output}_2=ReLU(Linear(\text{Input}_{64\xrightarrow{}32}))
    $$
    Layer 3: A final linear layer that takes the 32-dimensional output and maps it to a single-dimensional latent representation. No activation function is applied to this final output.
    $$\text{Latent Representation}=Linear(\text{Input}_{32\xrightarrow{}1})$$
\paragraph{Decoder}
The decoder takes the single-dimensional latent representation from the encoder and reconstructs the original 2-dimensional data. It mirrors the structure of the encoder, consisting of three fully connected layers with ReLU activations on the first two. The layers are structured as follows \\ \\ Layer 1: A linear layer that takes the 1-dimensional latent representation and outputs a 32-dimensional tensor. This is followed by a ReLU activation function.
    $$
    \text{Output}_1=ReLU(Linear(\text{Input}_{1\xrightarrow{}32}))
    $$
    Layer 2: A linear layer that takes the 32-dimensional output from the previous layer and outputs a 64-dimensional tensor. This is also followed by a ReLU activation function.
    $$
    \text{Output}_1=ReLU(Linear(\text{Input}_{32\xrightarrow{}64}))
    $$
    Layer 3: A final linear layer that takes the 64-dimensional output and reconstructs the 2-dimensional original data. No activation function is applied here, as the output represents the reconstructed values.
    $$\text{Reconstructed Data}=Linear(\text{Input}_{64\xrightarrow{}2})$$

\subsection{The Training Process: Putting the Model to Work}
Once our model and data were ready, we began the training process. The goal was to train the autoencoder to minimize the difference between its reconstructed output and the original input data.
$$\text{Reconstructed Data}=\text{Decoder}(\text{Encoder}(\text{Input Data}))$$

\subsubsection{Hyperparameters}
We used the Adam \cite{Adam} optimizer to adjust our model's weights and biases during training. The primary objective was to minimize the Mean Squared Error (MSE) \cite{MSE}, a standard loss function for regression tasks, which quantifies the reconstruction error.
$$\text{MSE}=\frac{1}{n}\sum_{i=1}^n(y_i - \hat{y}_i)^2$$
We trained the model for 1000 epochs, which was sufficient for the model's performance to stabilize. A crucial part of our study was exploring the impact of the learning rate, a hyperparameter that controls how quickly the model updates its weights. We conducted experiments using three distinct learning rates to observe their effects on the model's learning dynamics: 0.01, 0.001, and 0.0001.
\subsubsection{Monitoring the Model's Internal State}
A core component of our methodology was our ability to capture detailed data from inside the network during training. To achieve this, we developed a custom hooking mechanism in PyTorch. This mechanism acts like a data tap, capturing a wealth of information from each layer at the end of every training epoch.\\
We captured everything from weights and biases to activations and their gradients. To ensure we had a complete picture, we also captured standardized versions of this data. This comprehensive approach allowed us to see not just what the model was doing, but how it was learning, providing an in-depth view of the training dynamics.\\
The captured data was then organized and stored using the ROOT \cite{ROOT} data analysis framework. This choice was essential for handling the massive amount of time-series data generated, as ROOT provides a memory-efficient and structured way to manage and access large datasets.
\subsubsection{Visualizing and Analyzing the Results}
The training was performed on a CPU and took approximately two hours per learning rate. After training, we used the Matplotlib \cite{Matplotlib} library to visualize the results. These plots, which compare the original dataset with the reconstructed output, provide a clear visual representation of the autoencoder's performance and the impact of the different learning rates. The plots and a detailed discussion of our findings are presented in the following sections.

\section{Results}
We have first presented the reconstructions results for each learning rate on the spiral dataset and then we have presented the "spread of the spread" of five parameters during the training of our Autoencoder to see how the fluctuations occur inside both the decoder and encoder. We have hist-plots to visualize the fluctuations and the corresponding fluctuation table to show the raw figures. We have used matplotlib \cite{Matplotlib} to plot both the hist-plots and tables.
\subsection{Reconstruction Results}
\begin{figure}[H]
    \centering
    \caption{Reconstruction with learning rate 0.01}
    \includegraphics[scale=0.6]{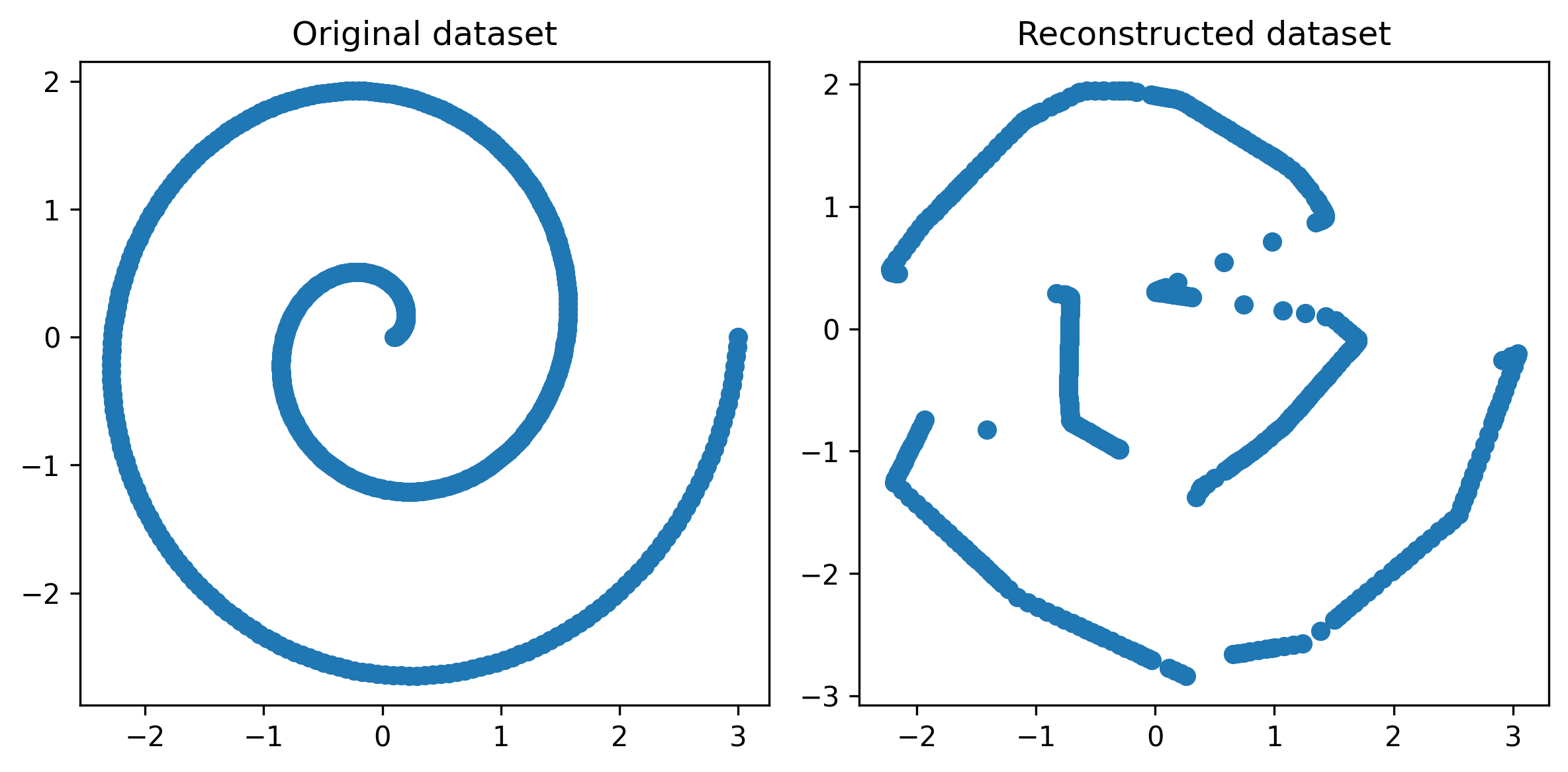}
    \label{fig:lr0.01}
\end{figure}
\begin{figure}[H]
    \centering
    \caption{Reconstruction with learning rate 0.001}
    \includegraphics[scale=0.6]{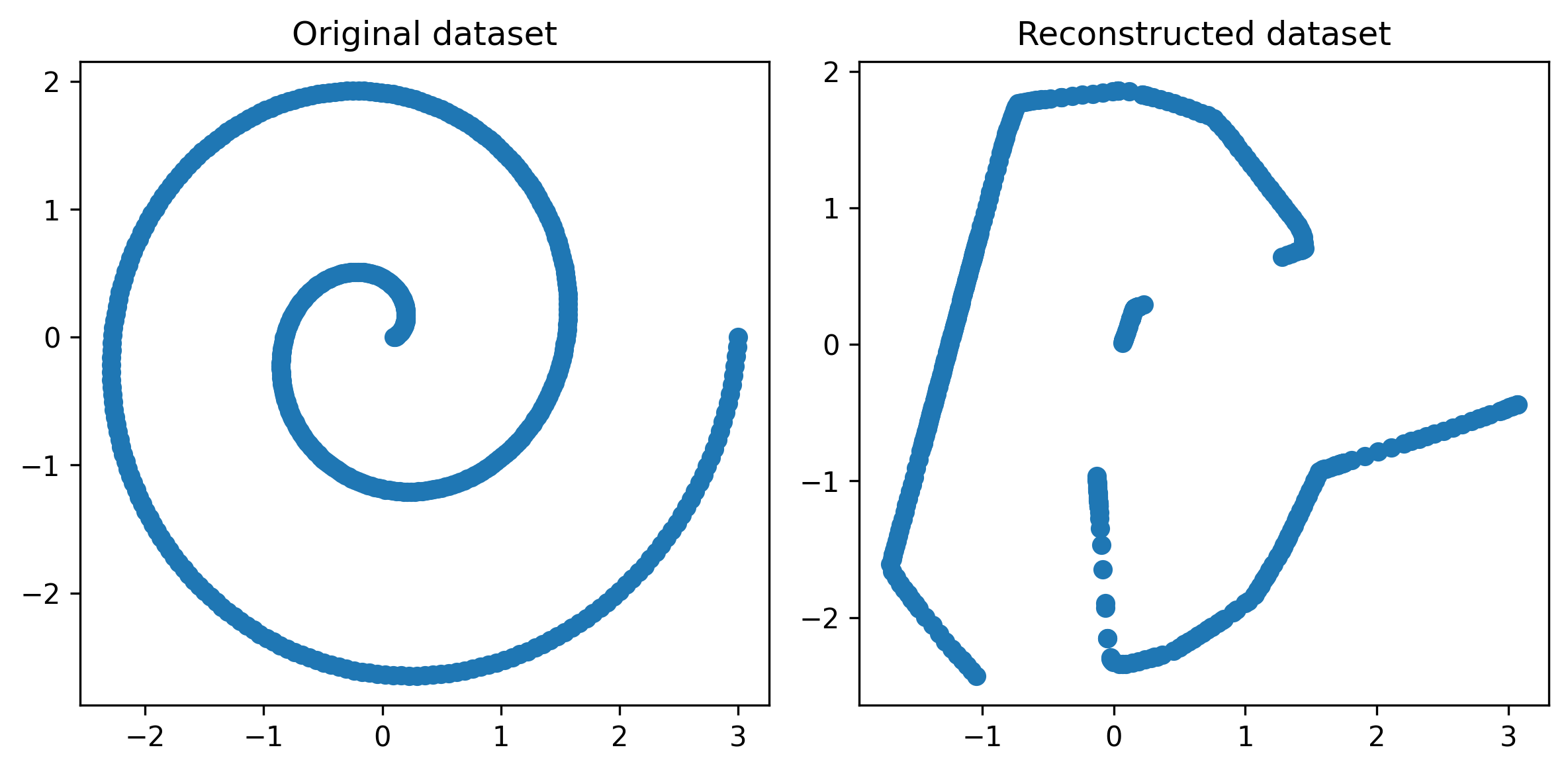}
    \label{fig:lr0.001}
\end{figure}
\begin{figure}[H]
    \centering
    \caption{Reconstruction with learning rate 0.0001}
    \includegraphics[scale=0.6]{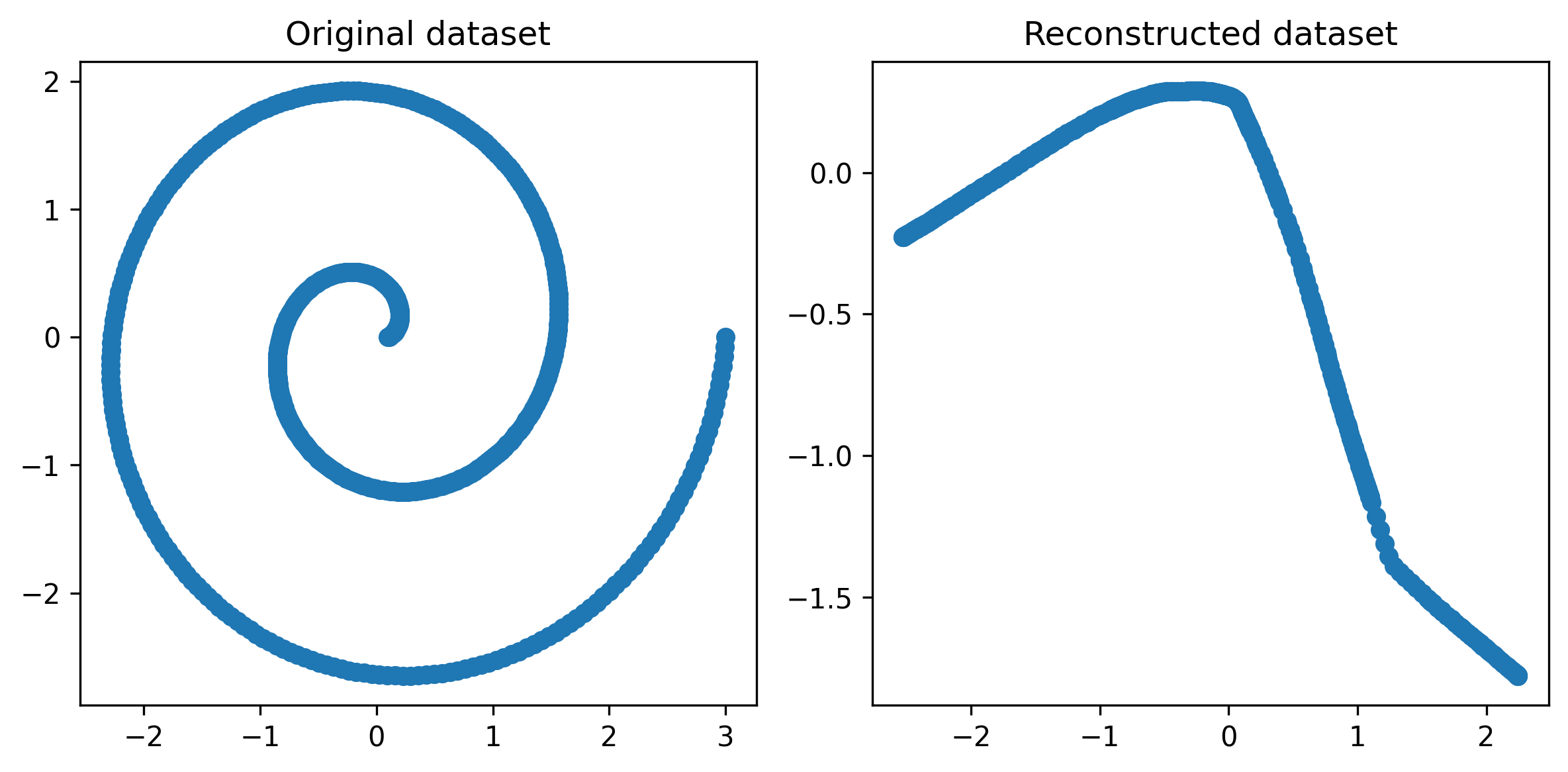}
    \label{fig:lr0.0001}
\end{figure}
\subsection{Fluctuations}
\begin{figure}[H]
    \centering
    \caption{Fluctuations in the weights of the model}
    \includegraphics[width=1\linewidth]{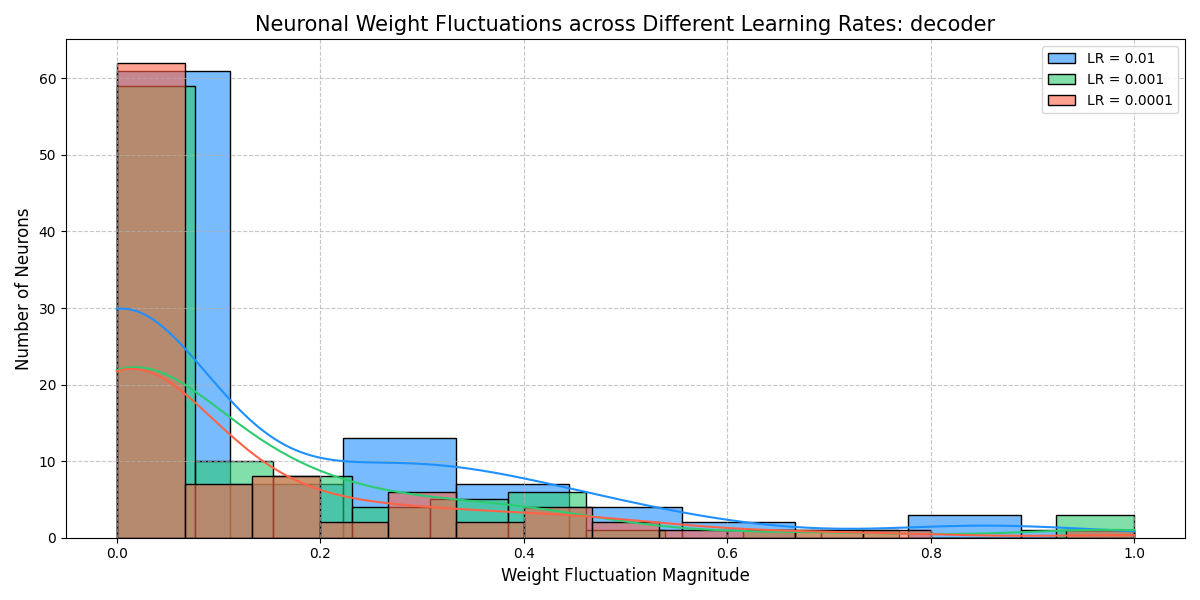}
    \includegraphics[width=1\linewidth]{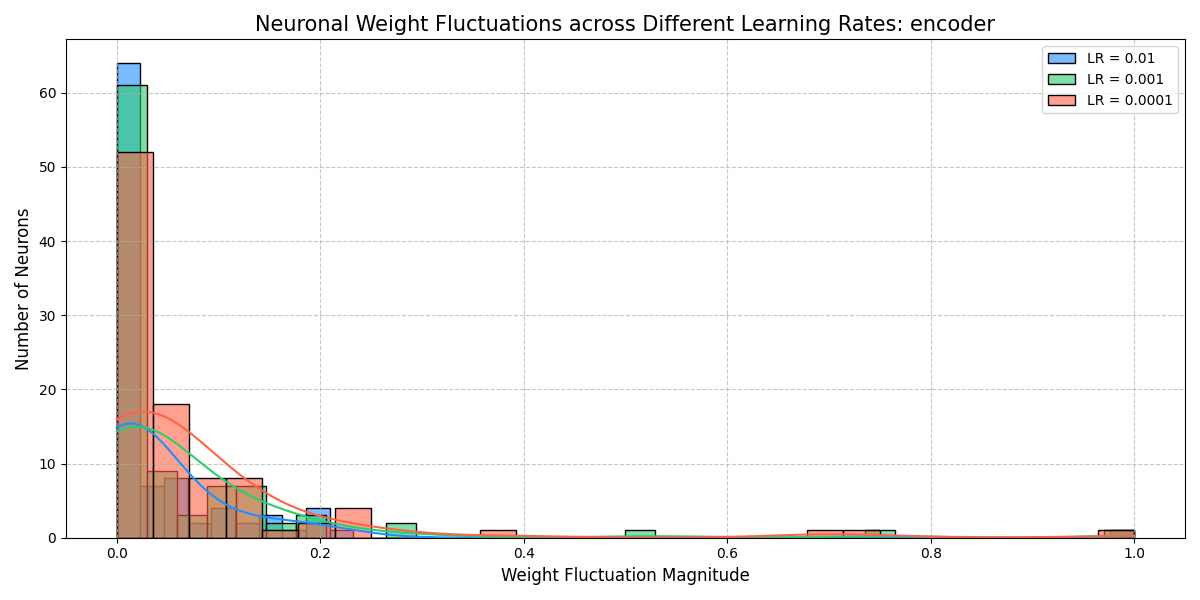}
    \includegraphics[width=1\linewidth]{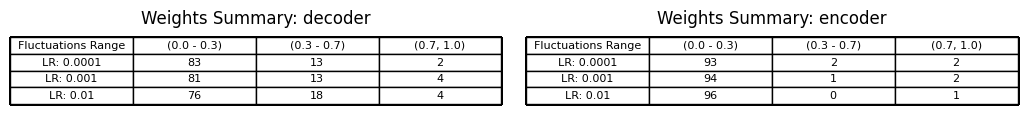}
    \label{fig:wf}
\end{figure}
\begin{figure}[H]
    \centering
    \caption{Fluctuations in the biases of the model}
    \includegraphics[width=1\linewidth]{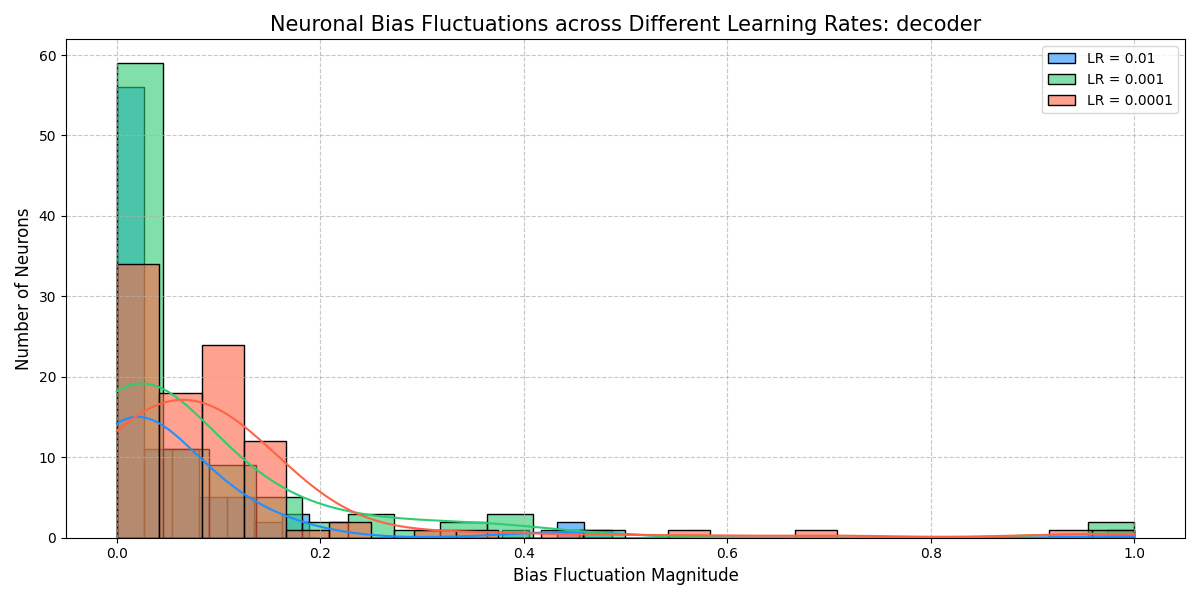}
    \includegraphics[width=1\linewidth]{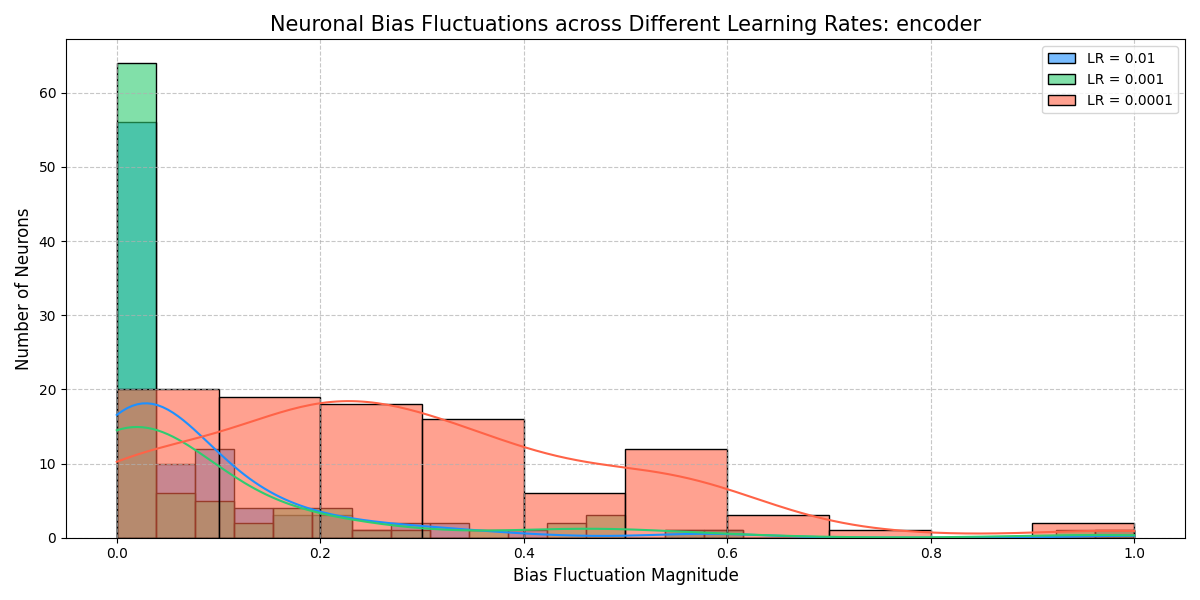}
    \includegraphics[width=1\linewidth]{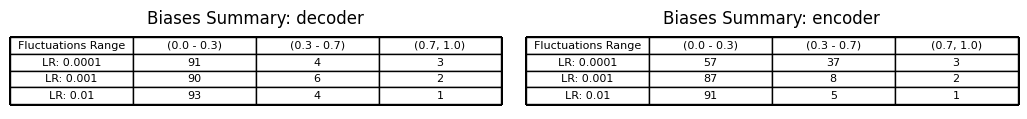}
    \label{fig:bf}
\end{figure}
\begin{figure}[H]
    \centering
    \caption{Fluctuations in the activations of the model}
    \includegraphics[width=1\linewidth]{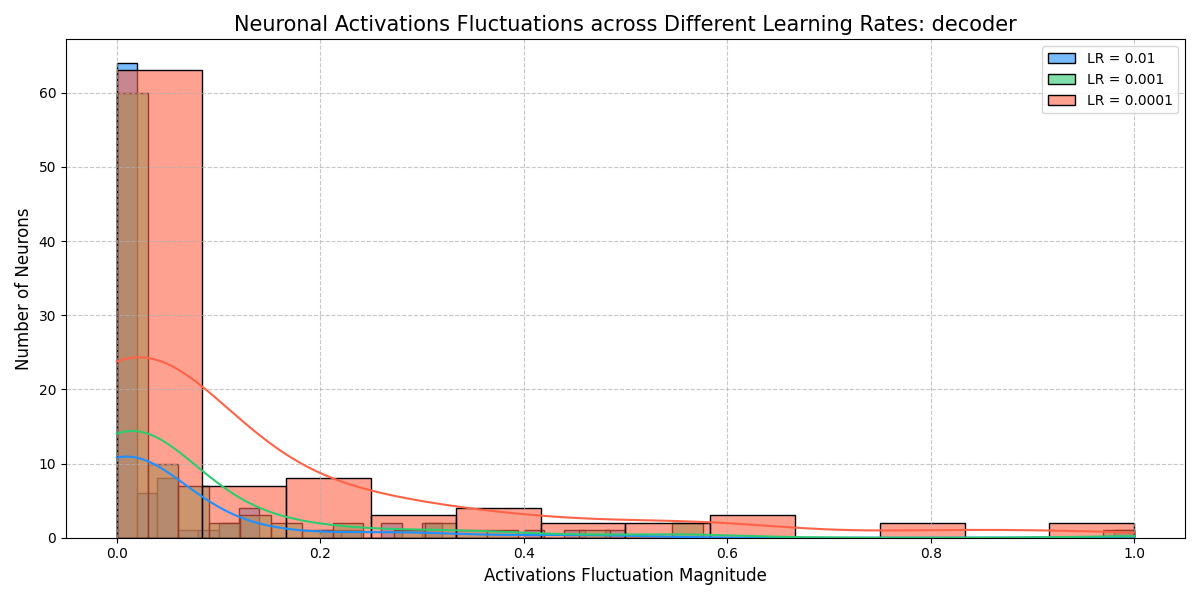}
    \includegraphics[width=1\linewidth]{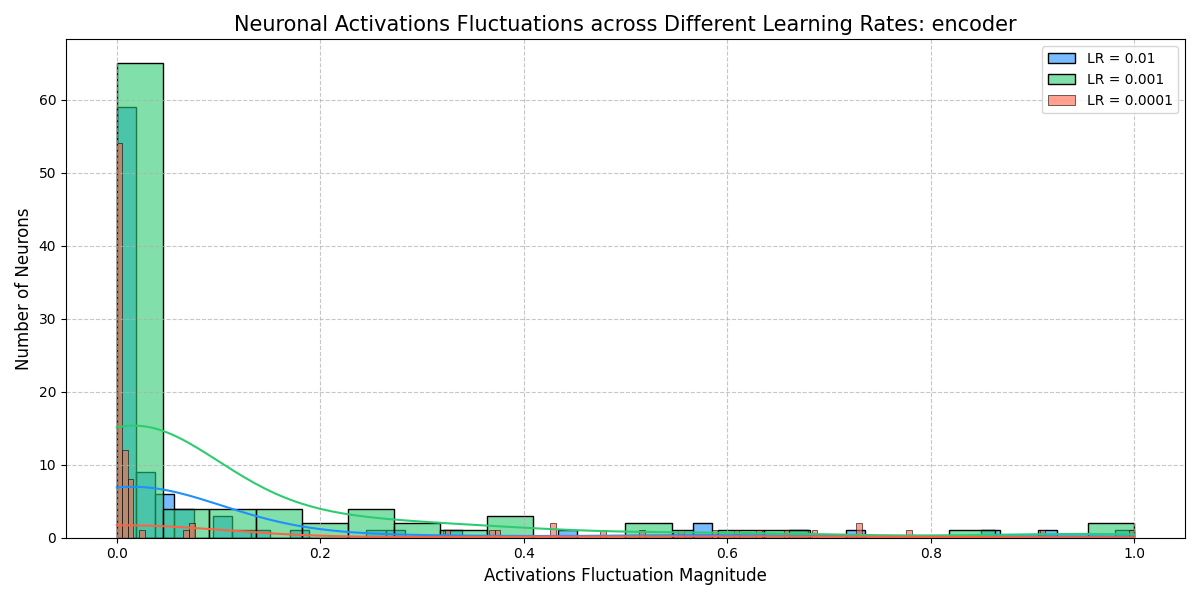}
    \includegraphics[width=1\linewidth]{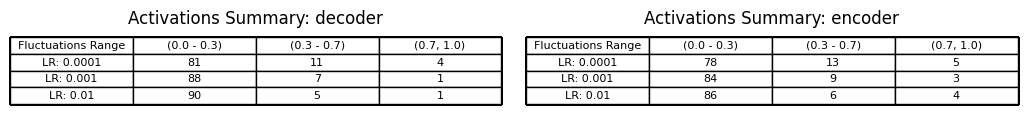}
    \label{fig:af}
\end{figure}
\begin{figure}[H]
    \centering
    \caption{Fluctuations in the weights of the model}
    \includegraphics[width=1\linewidth]{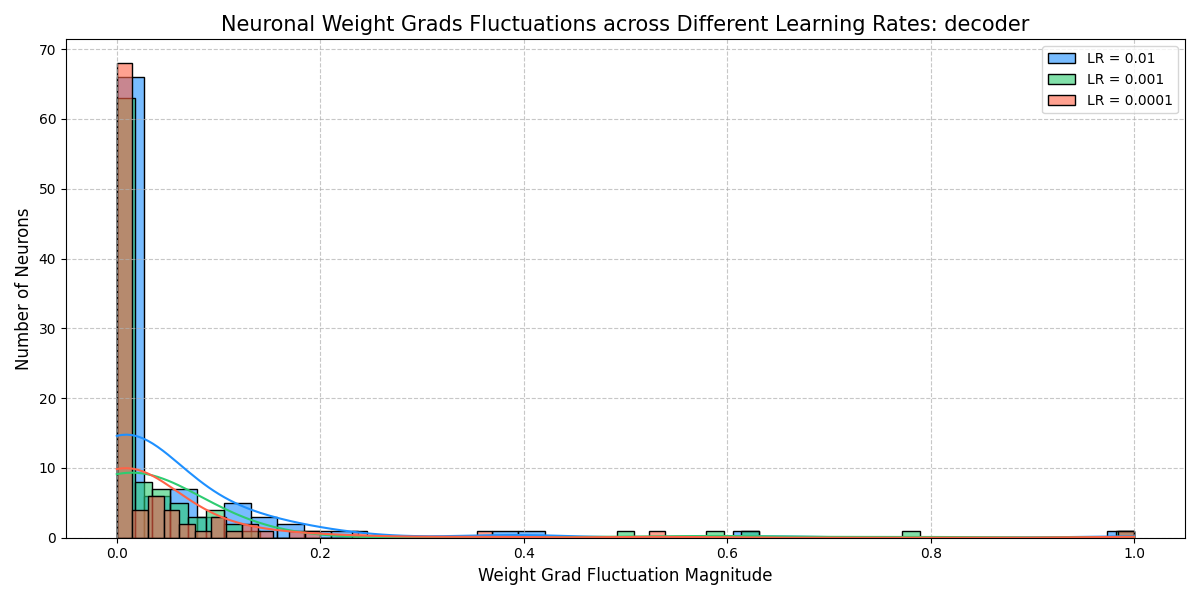}
    \includegraphics[width=1\linewidth]{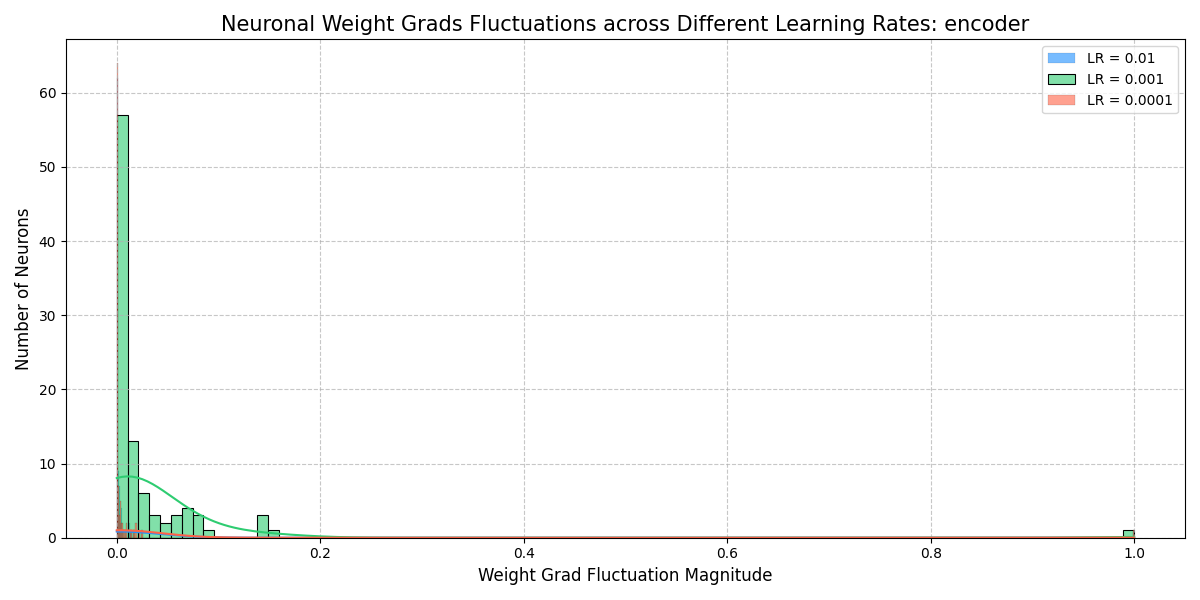}
    \includegraphics[width=1\linewidth]{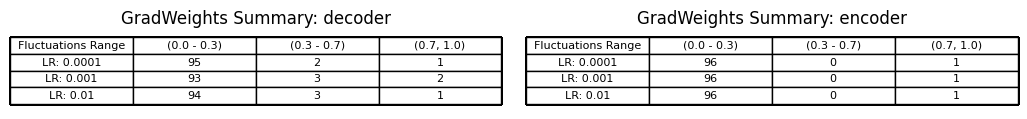}
    \label{fig:gwf}
\end{figure}
\begin{figure}[H]
    \centering
    \caption{Fluctuations in the weights of the model}
    \includegraphics[width=1\linewidth]{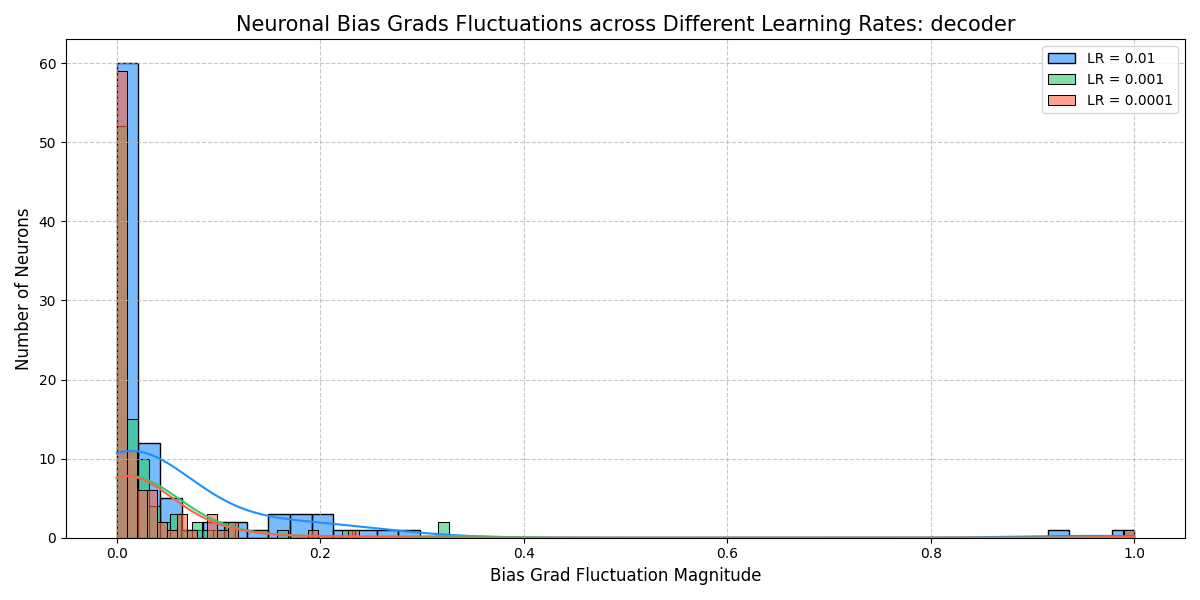}
    \includegraphics[width=1\linewidth]{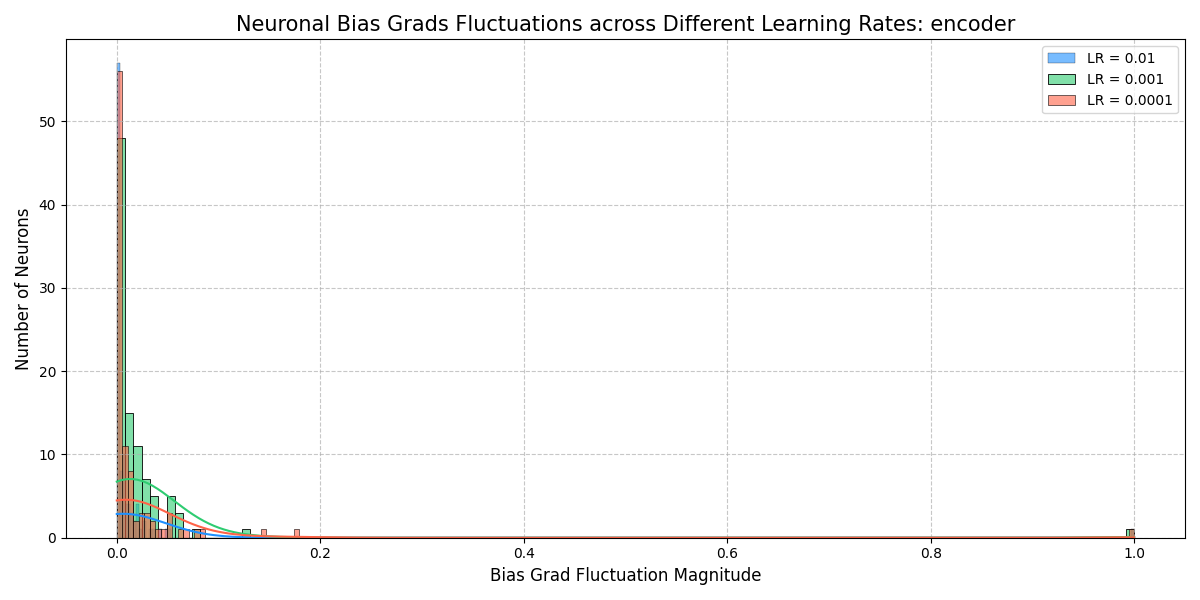}
    \includegraphics[width=1\linewidth]{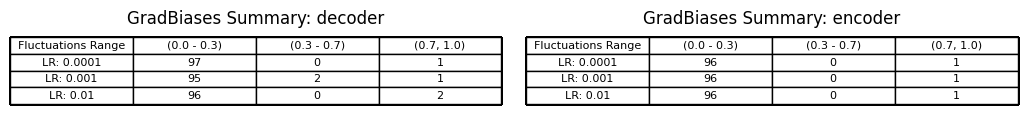}
    \label{fig:gbf}
\end{figure}
\section{Discussion}
In this section we will interpret the results from the previous section and try to reach to meaningful conclusions.
\subsection{Reconstruction Results}
From Figure \ref{fig:lr0.01}, the model demonstrates an understanding of the dataset's circular nature, attempting to reconstruct it. However, the reconstructed parts are notably fragmented. Despite this, it represents the best reconstruction among all three learning rates tested. Extending the training duration would likely improve the reconstruction quality.

Figure \ref{fig:lr0.001} shows that with a learning rate of $0.001$, the model is less successful at reconstruction compared to the $0.01$ learning rate. This is likely due to the slower learning process, which requires more training time to achieve a comparable result. However, a key observation is that the reconstructed parts are less fragmented and appear smoother.

Figure \ref{fig:lr0.0001} reveals that with a learning rate of $0.0001$, there are no discernible signs of reconstruction, a trend observed across most datasets we tested. This is attributed to the very low learning rate. While more training time would likely lead to better reconstruction, it's worth noting the absence of fragmentation, resulting in a very smooth curve.
\subsection{Fluctuations Results}
All the fluctuations are calculated as the "spread of the spread" which is a scalar metric that quantifies the dispersion of per-neuron parameter-change distributions across a neural network. This metric serves as a measure of the network's structural and dynamic consistency. This gave us a deep insight into the changes that are happening with in the model, most importantly we can identify the number of neurons whose parameters experienced close to zero fluctuations and we can then say that these neurons experienced no learning at all.

\subsubsection{Weight Fluctuations}
We can clearly see in the figure \ref{fig:wf} that a large number of neurons $ > 50 $ experienced close to $0$ fluctuation in the weights during the training of both encoder and decoder, this is also clear from the table under the plots.

We can also see that for the decoder the learning rate of $0.01$ showed the learning in most neurons throughout the different ranges but it is not the same in the case of the encoder. This can mean that the encoder of the model needs to be more finely tuned than the decoder.

\subsubsection{Bias Fluctuations}
We can see in the figure \ref{fig:bf} that again a large number of neurons experienced no change in their bias parameter. The decoder part of the model barely experienced any fluctuations in the bias terms across the learning rates. The encoder part of the model however shows that for the learning rate $0.0001$ there were massive fluctuations in the biases which are not present in the other two learning rates.

\subsubsection{Activation Fluctuations}
We can see in the plots of figure \ref{fig:af} that again maximum neurons experience very less fluctuations in this range, this is mainly because activations are directly related to respective neuron's weight and bias. We can also see clearly that learning rate of $0.0001$ is the best as it keeps the maximum number of neurons engaged in learning.
\paragraph{Scanning for Optimal Activation function range}
The learning rate and activation function act as a coupled system in a neural network; the optimal setting for one is highly dependent on the choice of the other \cite{lrcite}. Successfully training a model requires navigating the trade-offs inherent in this relationship.

For instance, traditional activation functions like sigmoid and tanh can saturate—that is, their outputs flatten out at the extremes. This causes the gradients that drive learning to shrink to almost nothing, effectively stalling the training process in what is known as the vanishing gradient problem.

The Rectified Linear Unit (ReLU) was developed to solve this, but it introduces a different failure mode. An aggressive learning rate can push a ReLU neuron's inputs into a negative state from which it can't recover. The neuron's output becomes permanently zero, preventing it from participating in learning—a problem called the "dying ReLU" phenomenon. Our study suggests that for a specific learning rate there exists a range of the activation function that is being used.

\subsubsection{Fluctuations in the Gradients}
As we see in the figures \ref{fig:gwf} and \ref{fig:gbf} both Bias and Weight gradients show similar fluctuations, most neurons have almost $0$ gradient and this is because the gradients of any neuron is already small, after multiplying it with the learning rates it becomes even more smaller, so this graph was expected. The plots confirm that the learning process is functioning as intended, with changes in weights and biases directly corresponding to the respective gradients, validating the model's behavior.

\subsection{Final Thoughts}
After analyzing the plots and figures we can draw the following conclusions:
\begin{itemize}
    \item Approximately $50$ neurons remained inactive throughout the $1000$-epoch training, showing minimal to no parameter change.
    \item A learning rate of $0.01$ yielded the best reconstruction but also resulted in the highest number of inactive neurons.
    \item The $0.001$ learning rate provided a comparable, though slightly inferior, reconstruction while showing a similar number of inactive neurons as the $0.01$ rate.
    \item Conversely, the $0.0001$ learning rate resulted in the poorest reconstruction, yet it was the most effective at engaging the majority of neurons in the learning process.
    \item This suggests that given a longer training duration, the $0.0001$ learning rate might have achieved a superior reconstruction, a hypothesis left for future investigation.
\end{itemize}

\subsection{Future Work}
Reaching the above conclusion we can think of the following ways of expanding this research.
\begin{itemize}
    \item Investigating the effect on accuracy by pruning the inactive neurons to understand if a smaller, more efficient network could achieve similar or better results.
    \item Analyzing the layer-wise impact of different learning rates to understand how various parts of the network learn and interact with each other.
    \item Studying the effect on power and resource consumption during both training and inference across different learning rates to evaluate the real-world efficiency of the model.
\end{itemize}

\printbibliography 
\appendix
\section{Results for other datasets}
\subsection{Triangle}
\begin{figure}[H]
    \centering
    \includegraphics[width=1\linewidth]{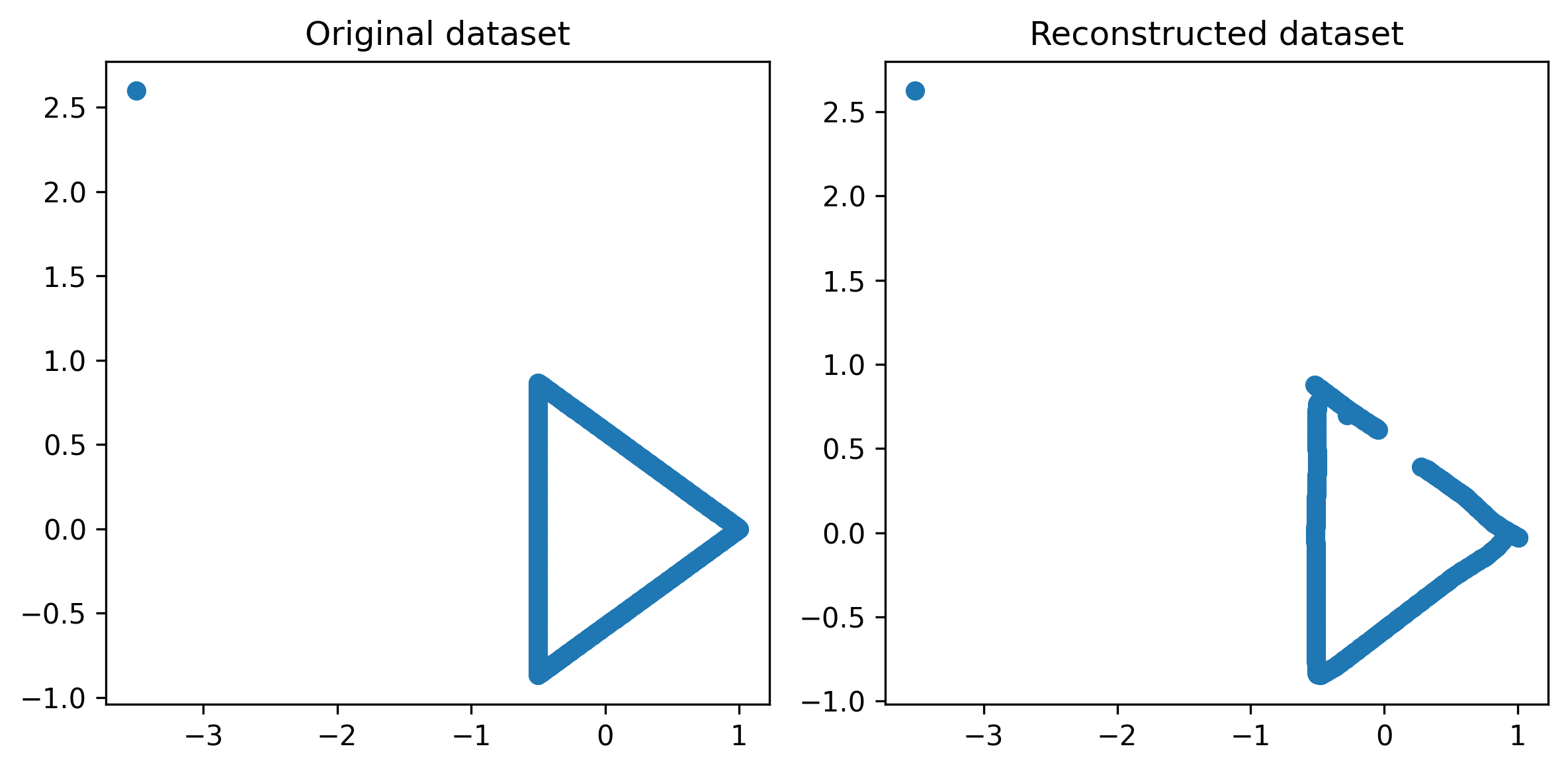}
    \caption{Recreation with learning rate 0.01}
\end{figure}
\begin{figure}[H]
    \includegraphics[width=1\linewidth]{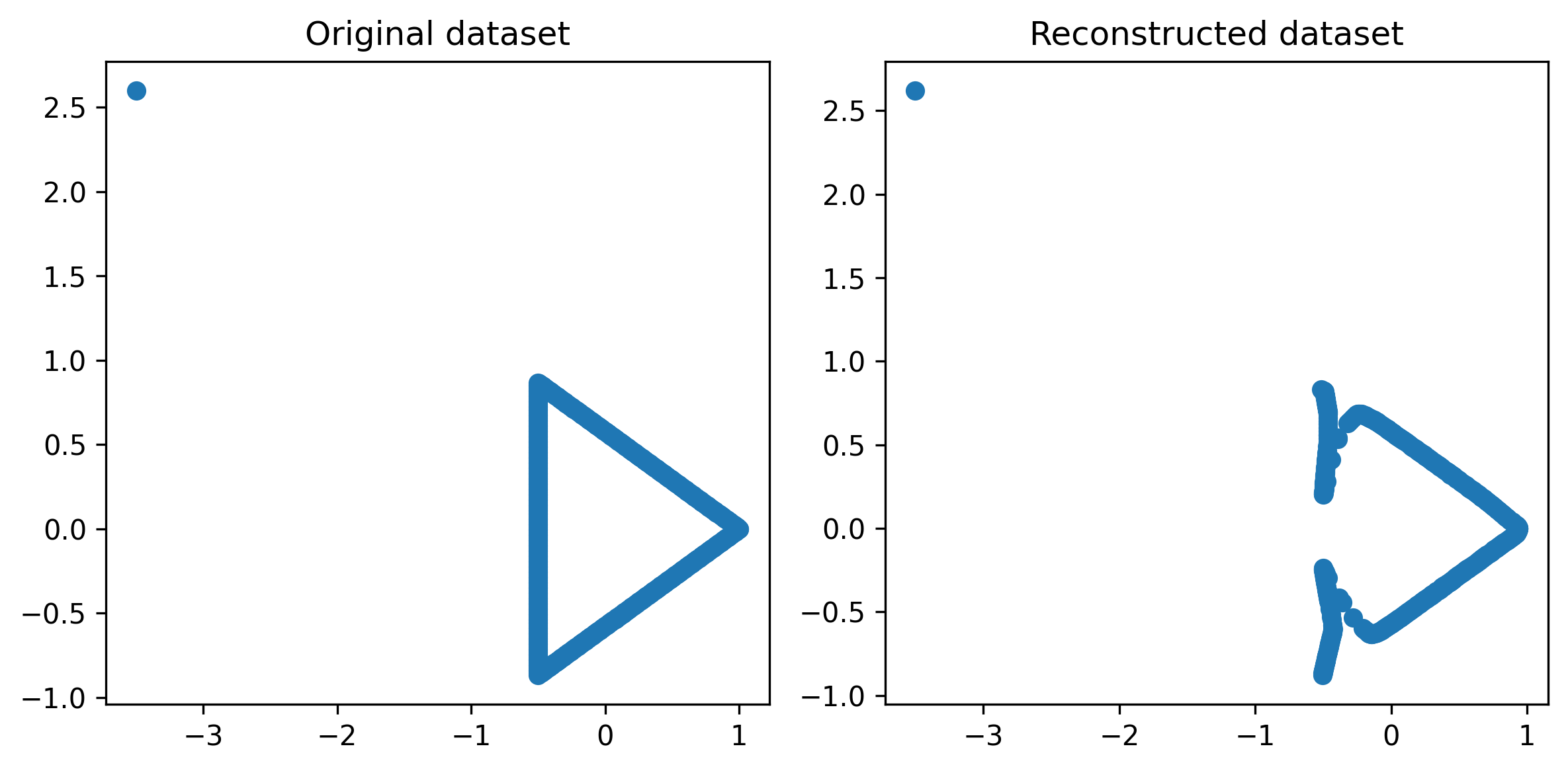}
    \caption{Recreation with learning rate 0.001}
\end{figure}
\begin{figure}[H]
    \includegraphics[width=1\linewidth]{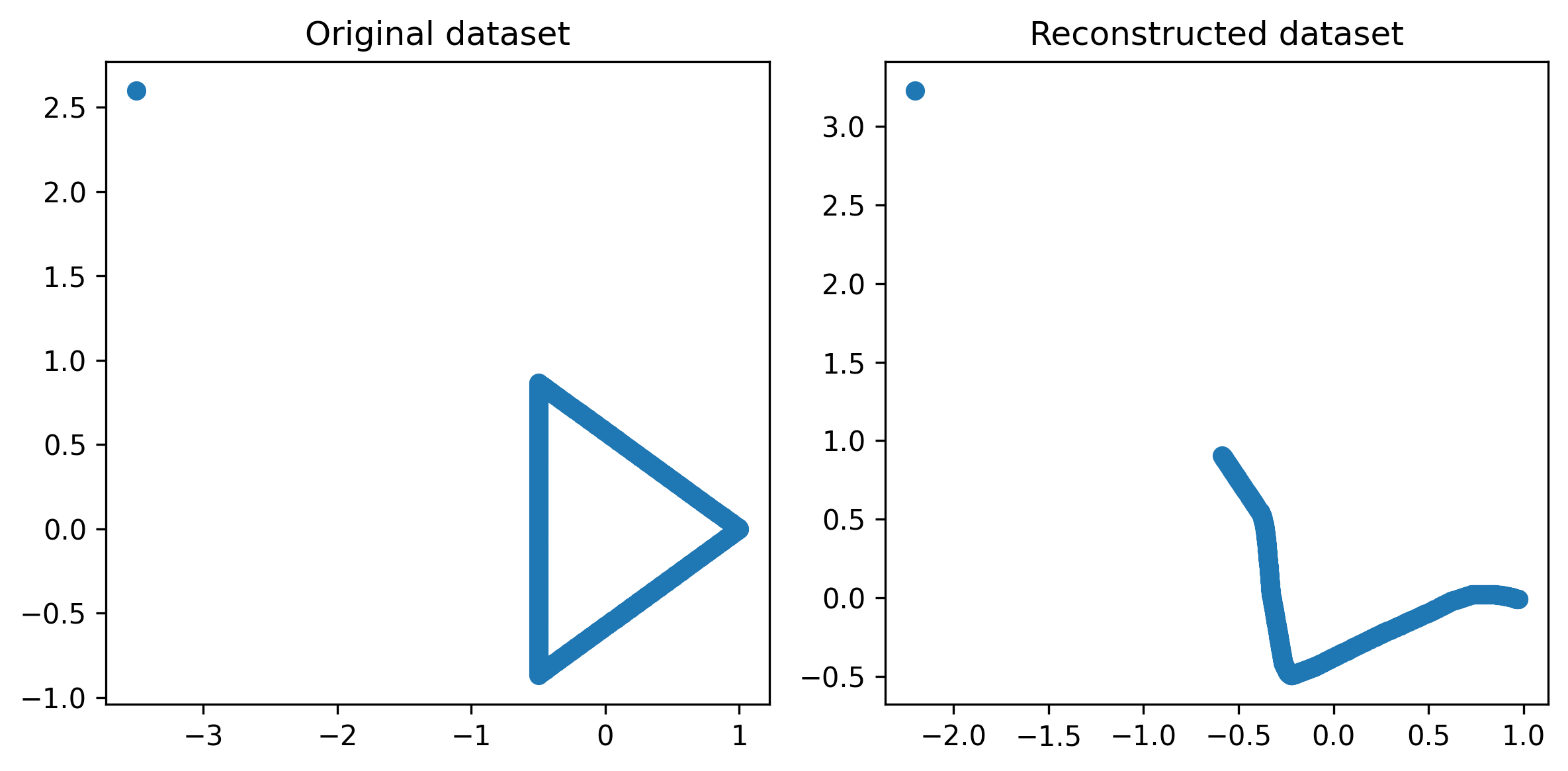}
    \caption{Recreation with learning rate 0.0001}
\end{figure}
\begin{figure}[H]
    \includegraphics[width=1\linewidth]{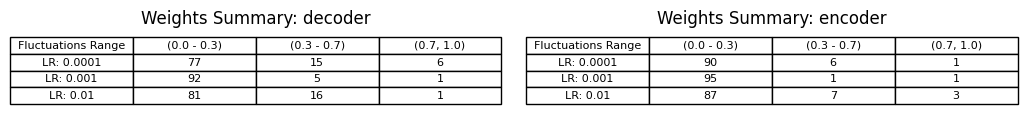}
    \caption{Fluctuations in Weights}
    \label{fig:TFW}
\end{figure}
\begin{figure}[H]
    \centering
    \includegraphics[width=1\linewidth]{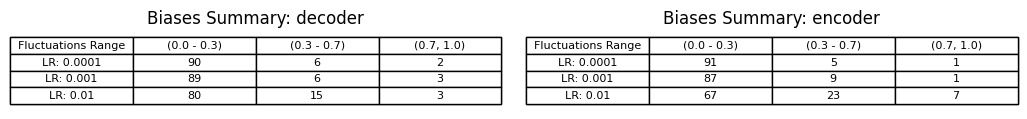}
    \caption{Fluctuations in Biases}
    \label{fig:TFB}
\end{figure}
\begin{figure}[H]
    \centering
    \includegraphics[width=1\linewidth]{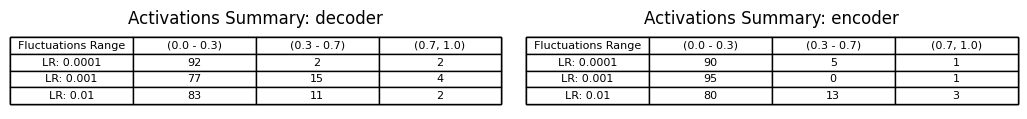}
    \caption{Fluctuations in Activations}
    \label{fig:TFA}
\end{figure}
\begin{figure}[H]
    \centering
    \includegraphics[width=1\linewidth]{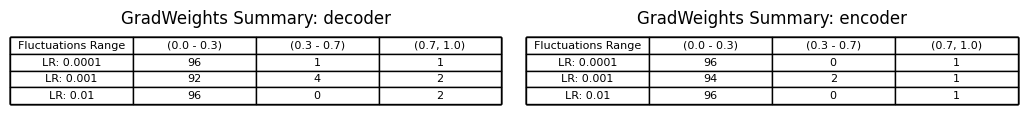}
    \caption{Fluctuations in Weight Gradient}
    \label{fig:TFGW}
\end{figure}
\begin{figure}[H]
    \centering
    \includegraphics[width=1\linewidth]{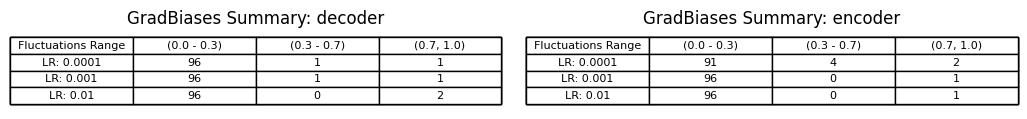}
    \caption{Fluctuations in Bias Gradient}
    \label{fig:TFGB}
\end{figure}
\subsection{Square}
\begin{figure}[H]
    \centering
    \includegraphics[width=1\linewidth]{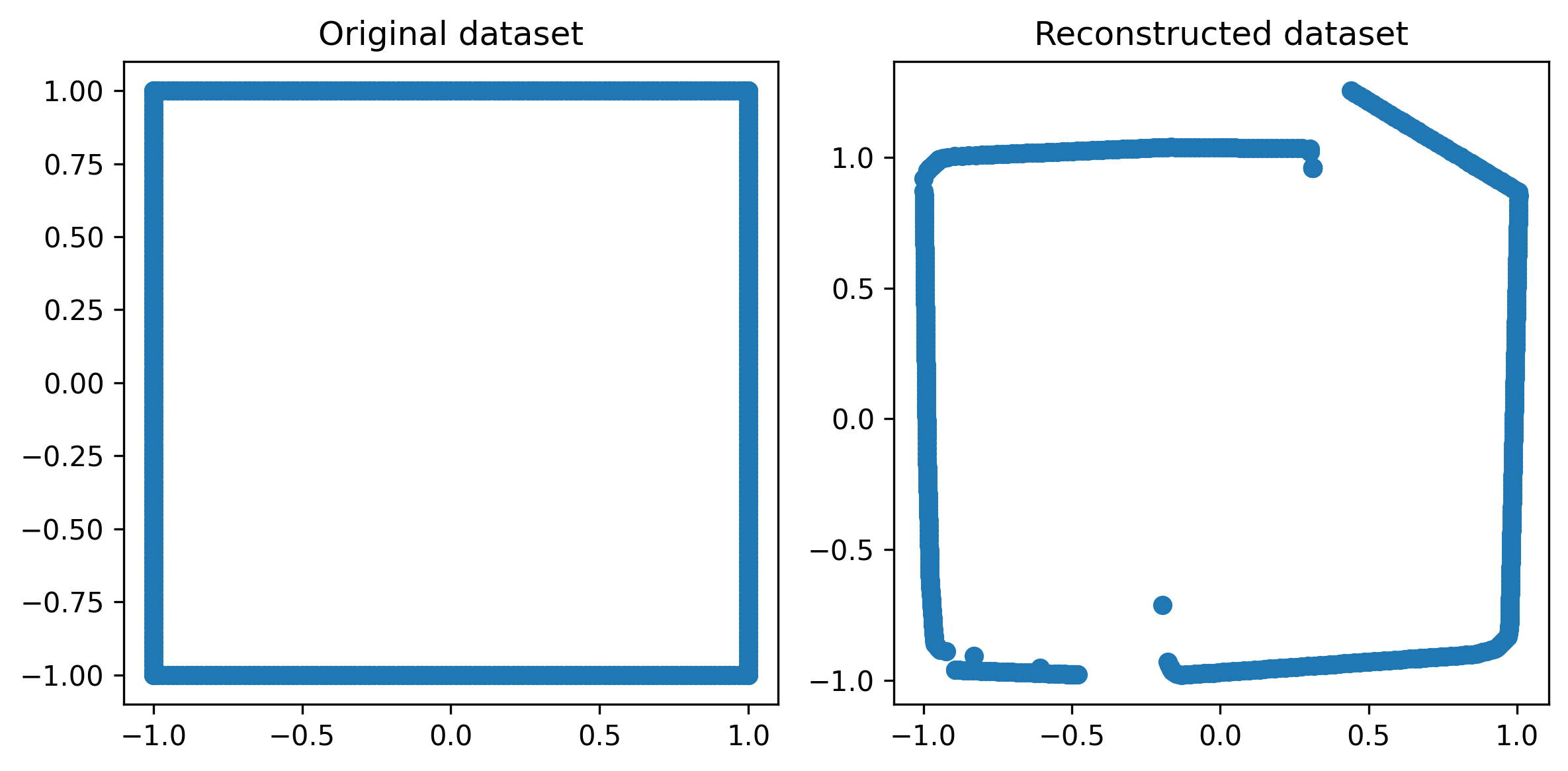}
    \caption{Recreation with learning rate 0.01}
\end{figure}
\begin{figure}[H]
    \includegraphics[width=1\linewidth]{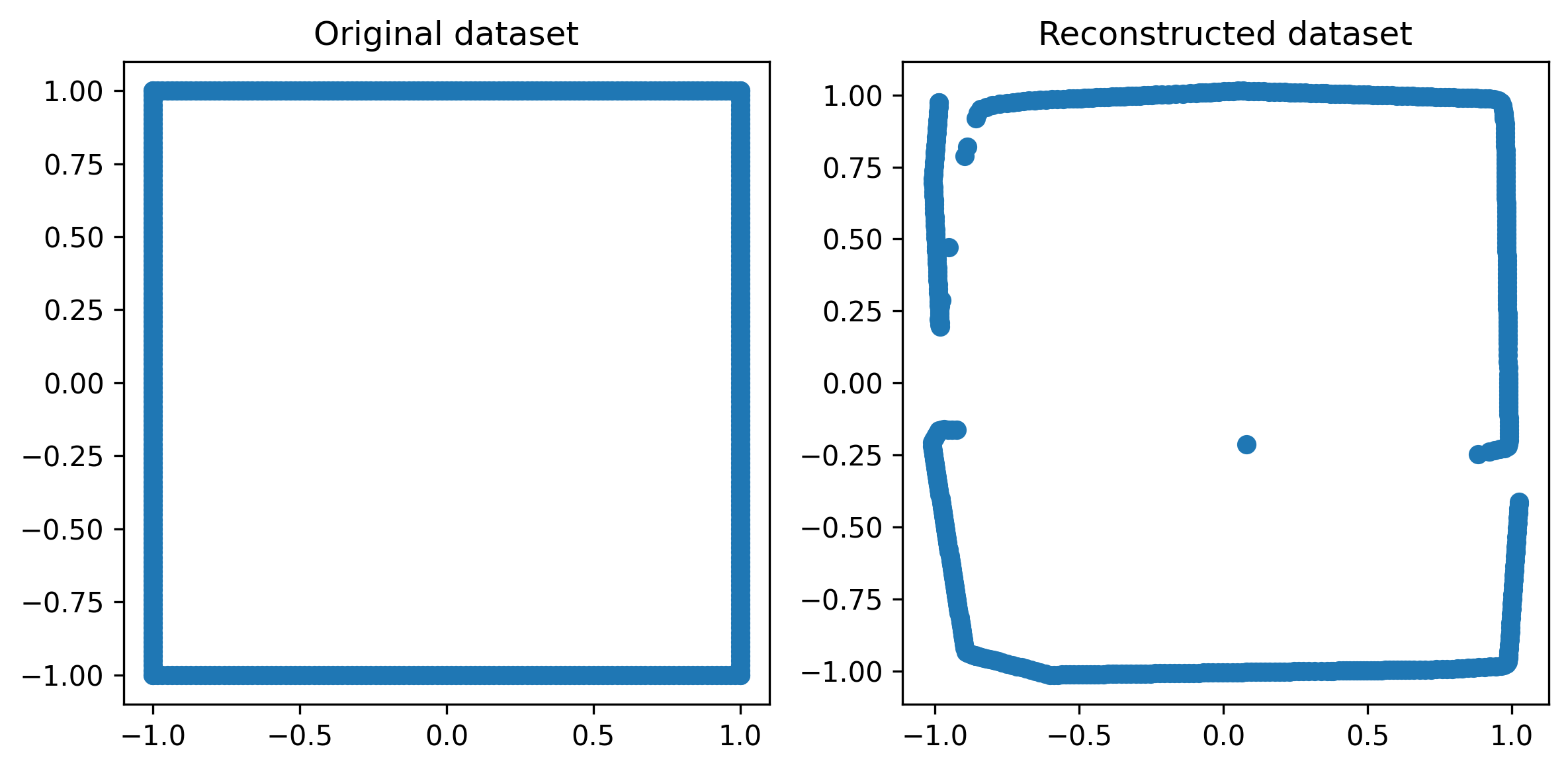}
    \caption{Recreation with learning rate 0.001}
\end{figure}
\begin{figure}[H]
    \includegraphics[width=1\linewidth]{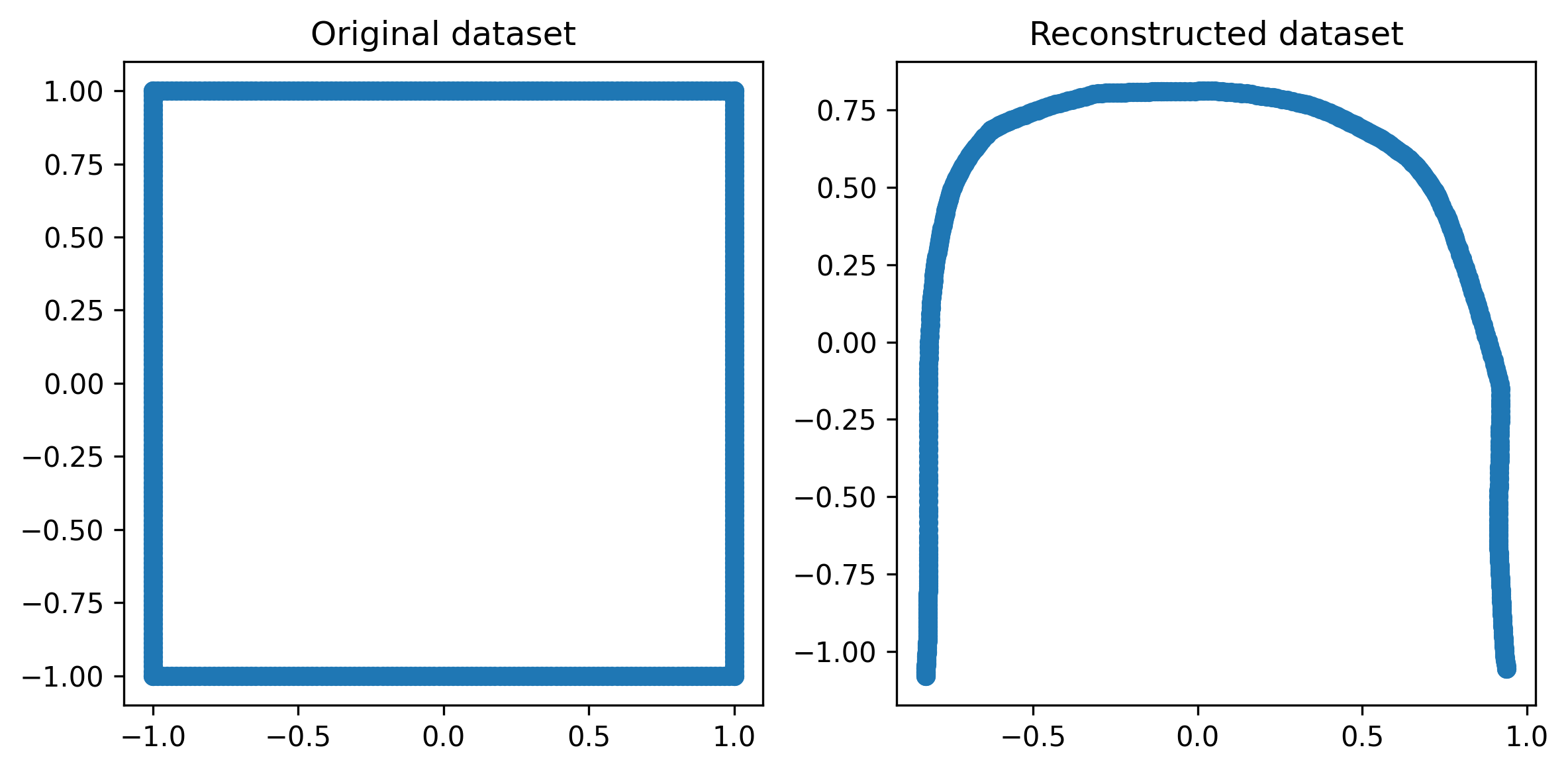}
    \caption{Recreation with learning rate 0.0001}
\end{figure}
\begin{figure}[H]
    \centering
    \includegraphics[width=1\linewidth]{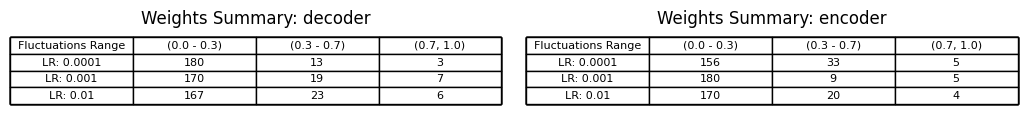}
    \caption{Fluctuations in Weights}
    \label{fig:SFW}
\end{figure}
\begin{figure}[H]
    \centering
    \includegraphics[width=1\linewidth]{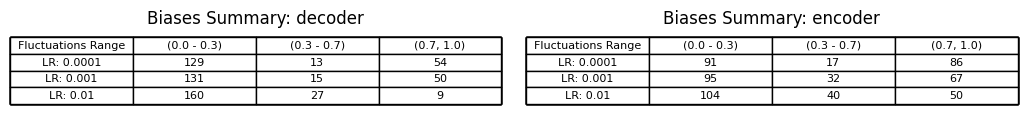}
    \caption{Fluctuations in Biases}
    \label{fig:SFB}
\end{figure}
\begin{figure}[H]
    \centering
    \includegraphics[width=1\linewidth]{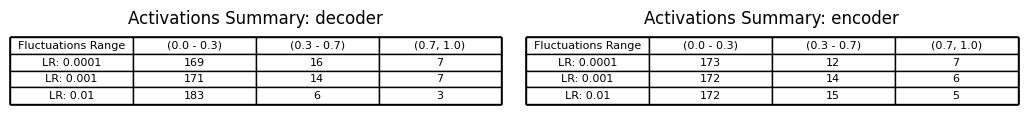}
    \caption{Fluctuations in Activations}
    \label{fig:SFA}
\end{figure}
\begin{figure}[H]
    \centering
    \includegraphics[width=1\linewidth]{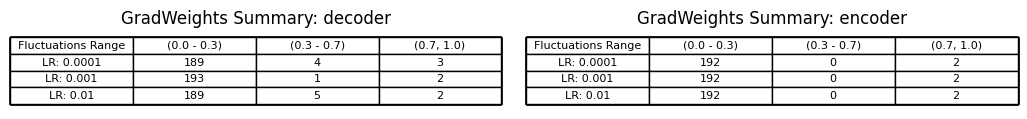}
    \caption{Fluctuations in Weight Gradient}
    \label{fig:SFGW}
\end{figure}
\begin{figure}[H]
    \centering
    \includegraphics[width=1\linewidth]{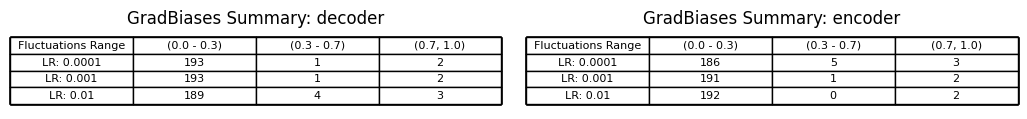}
    \caption{Fluctuations in Bias Gradient}
    \label{fig:SFGB}
\end{figure}
\subsection{Pentagon}
\begin{figure}[H]
    \centering
    \includegraphics[width=1\linewidth]{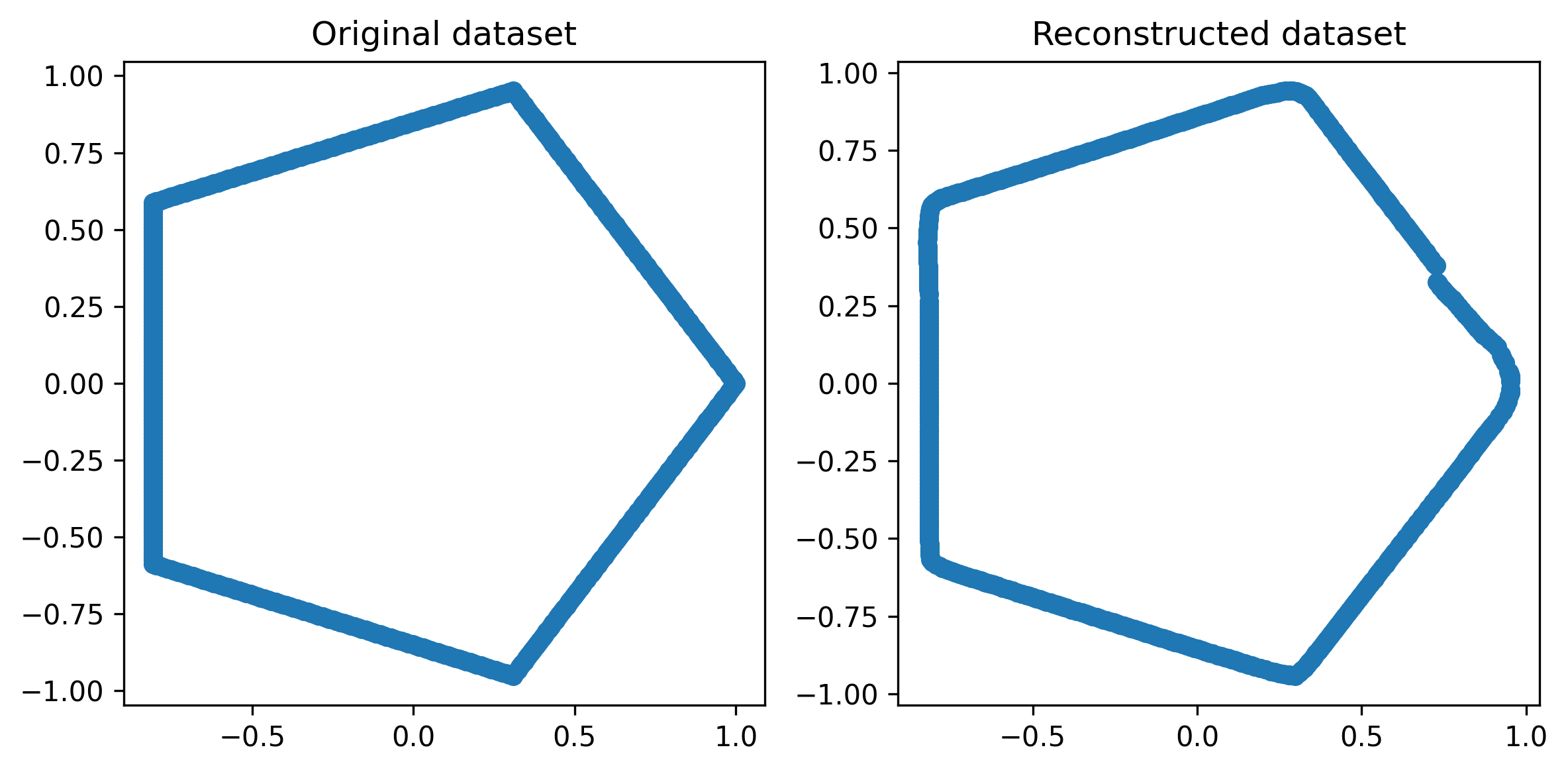}
    \caption{Recreation with learning rate 0.01}
\end{figure}
\begin{figure}[H]
    \includegraphics[width=1\linewidth]{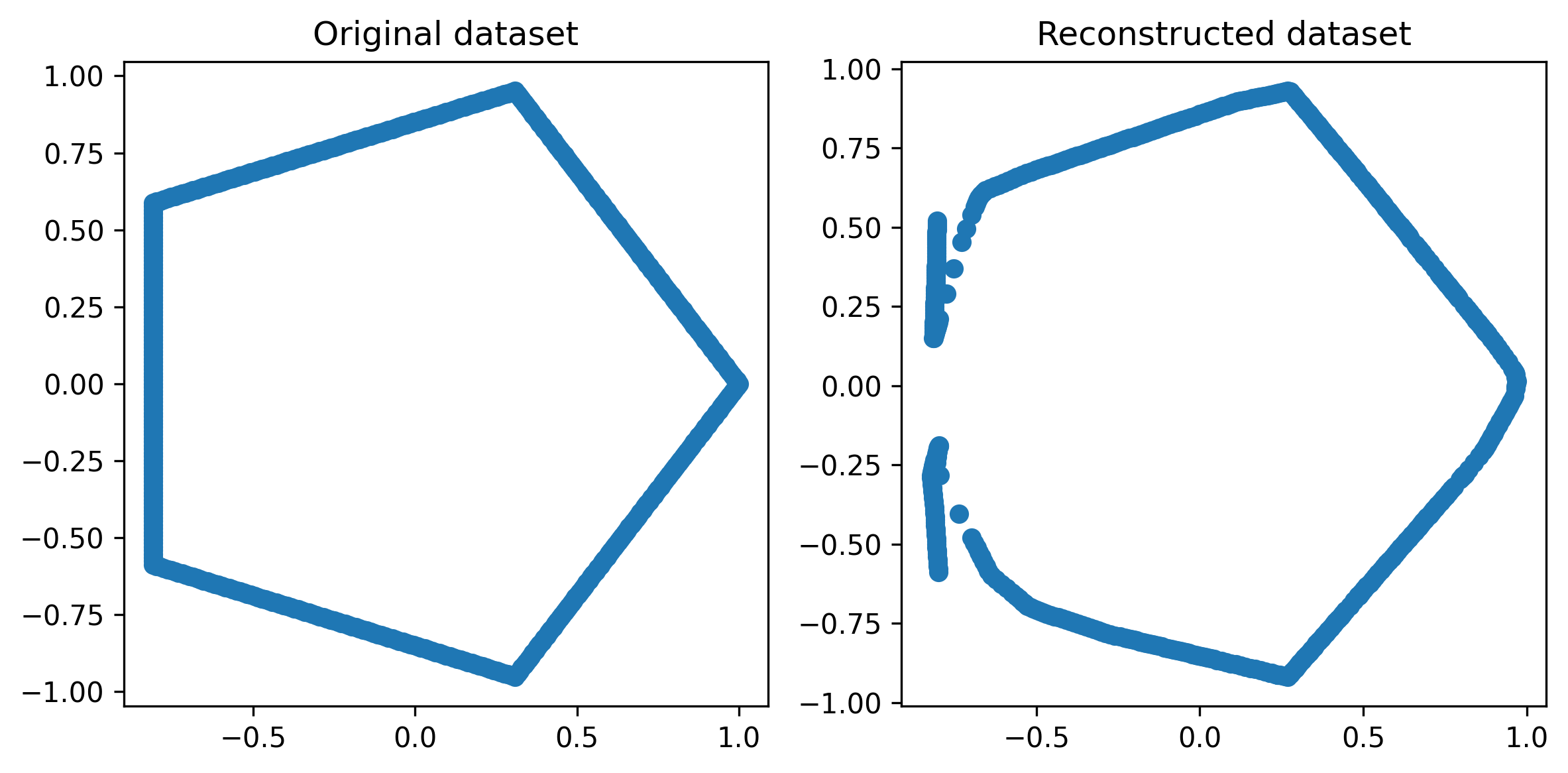}
    \caption{Recreation with learning rate 0.001}
\end{figure}
\begin{figure}[H]
    \includegraphics[width=1\linewidth]{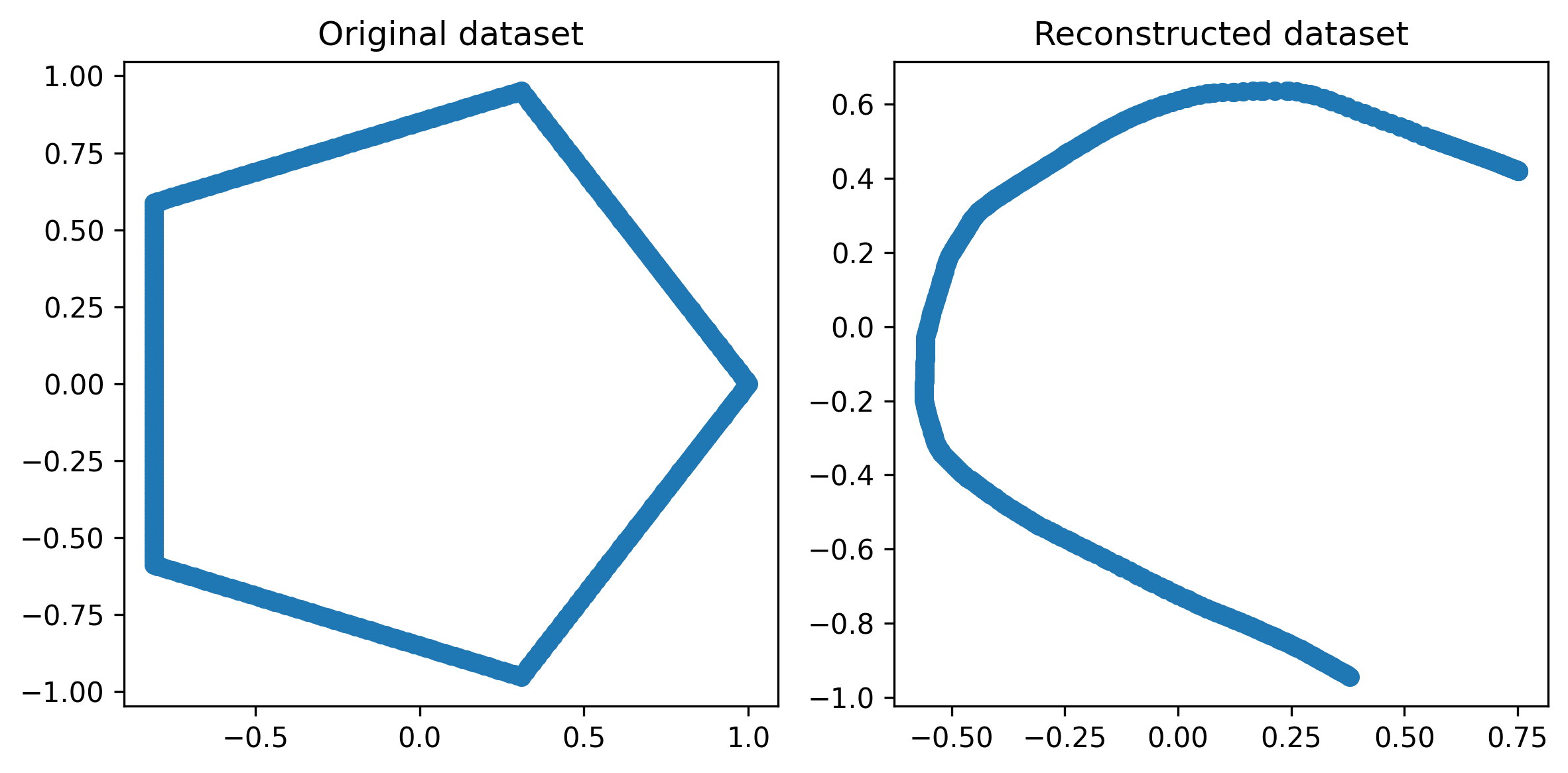}
    \caption{Recreation with learning rate 0.0001}
\end{figure}
\begin{figure}[H]
    \centering
    \includegraphics[width=1\linewidth]{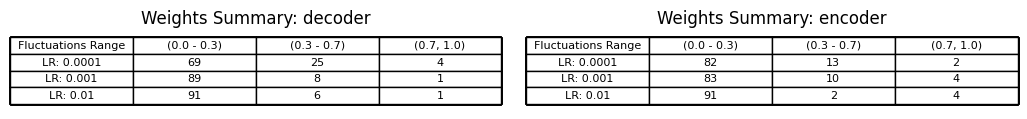}
    \caption{Fluctuations in Weights}
    \label{fig:PFW}
\end{figure}
\begin{figure}[H]
    \centering
    \includegraphics[width=1\linewidth]{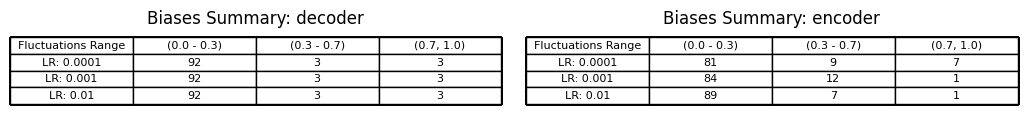}
    \caption{Fluctuations in Biases}
    \label{fig:PFB}
\end{figure}
\begin{figure}[H]
    \centering
    \includegraphics[width=1\linewidth]{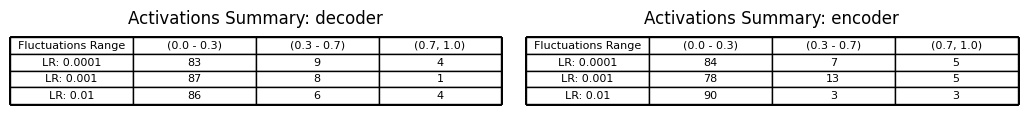}
    \caption{Fluctuations in Activations}
    \label{fig:PFA}
\end{figure}
\begin{figure}[H]
    \centering
    \includegraphics[width=1\linewidth]{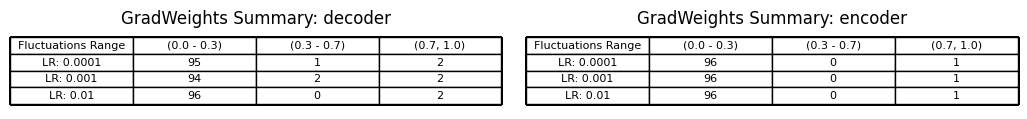}
    \caption{Fluctuations in Weight Gradient}
    \label{fig:PFGW}
\end{figure}
\begin{figure}[H]
    \centering
    \includegraphics[width=1\linewidth]{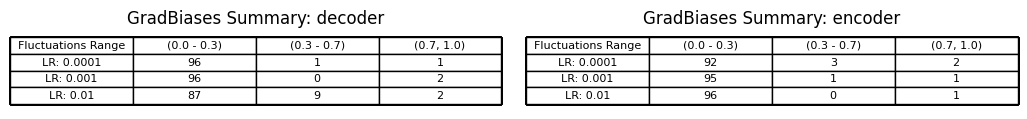}
    \caption{Fluctuations in Bias Gradient}
    \label{fig:PFGB}
\end{figure}
\subsection{Hexagon}
\begin{figure}[H]
    \centering
    \includegraphics[width=1\linewidth]{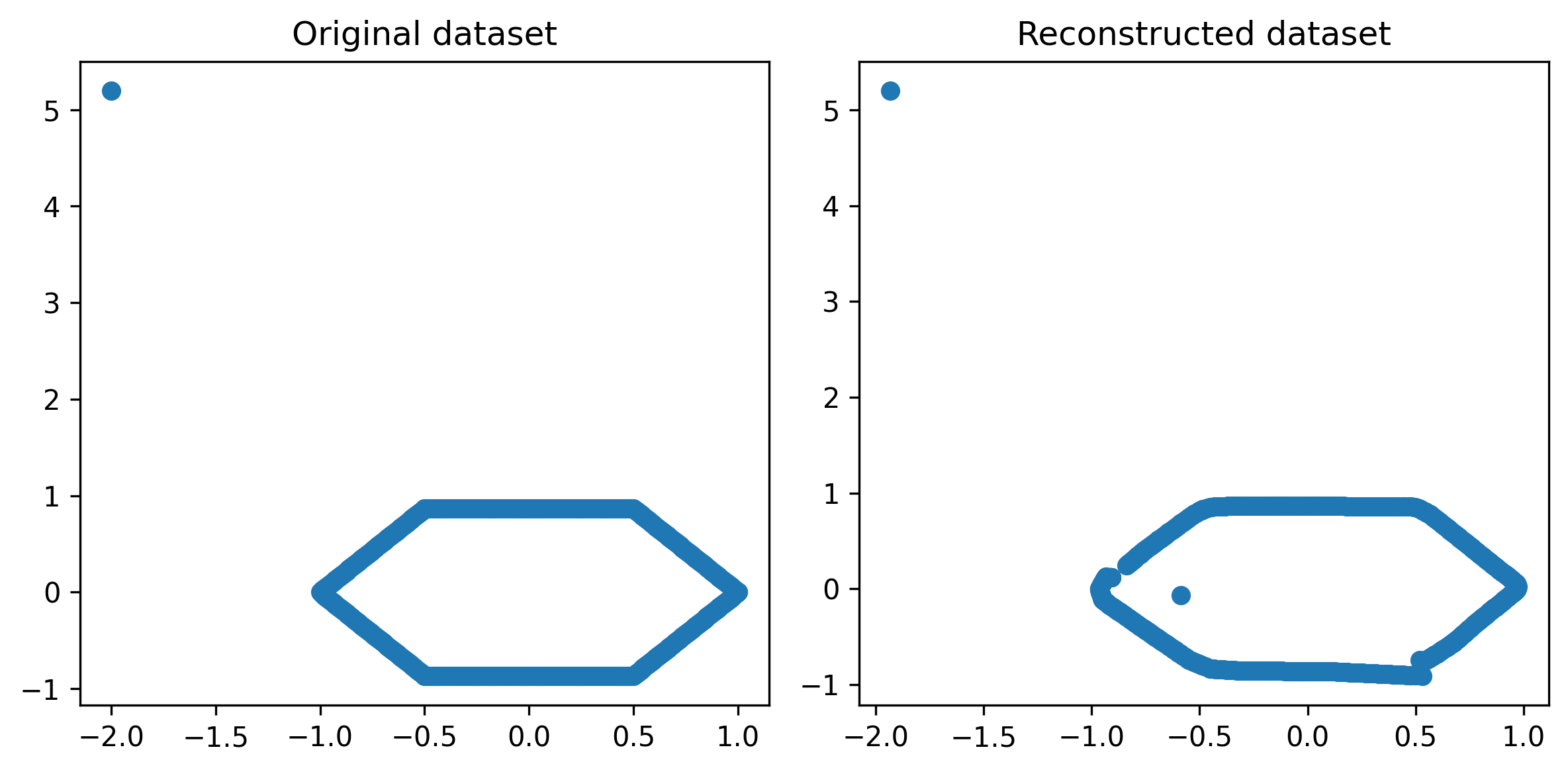}
    \caption{Recreation with learning rate 0.01}
\end{figure}
\begin{figure}[H]
    \includegraphics[width=1\linewidth]{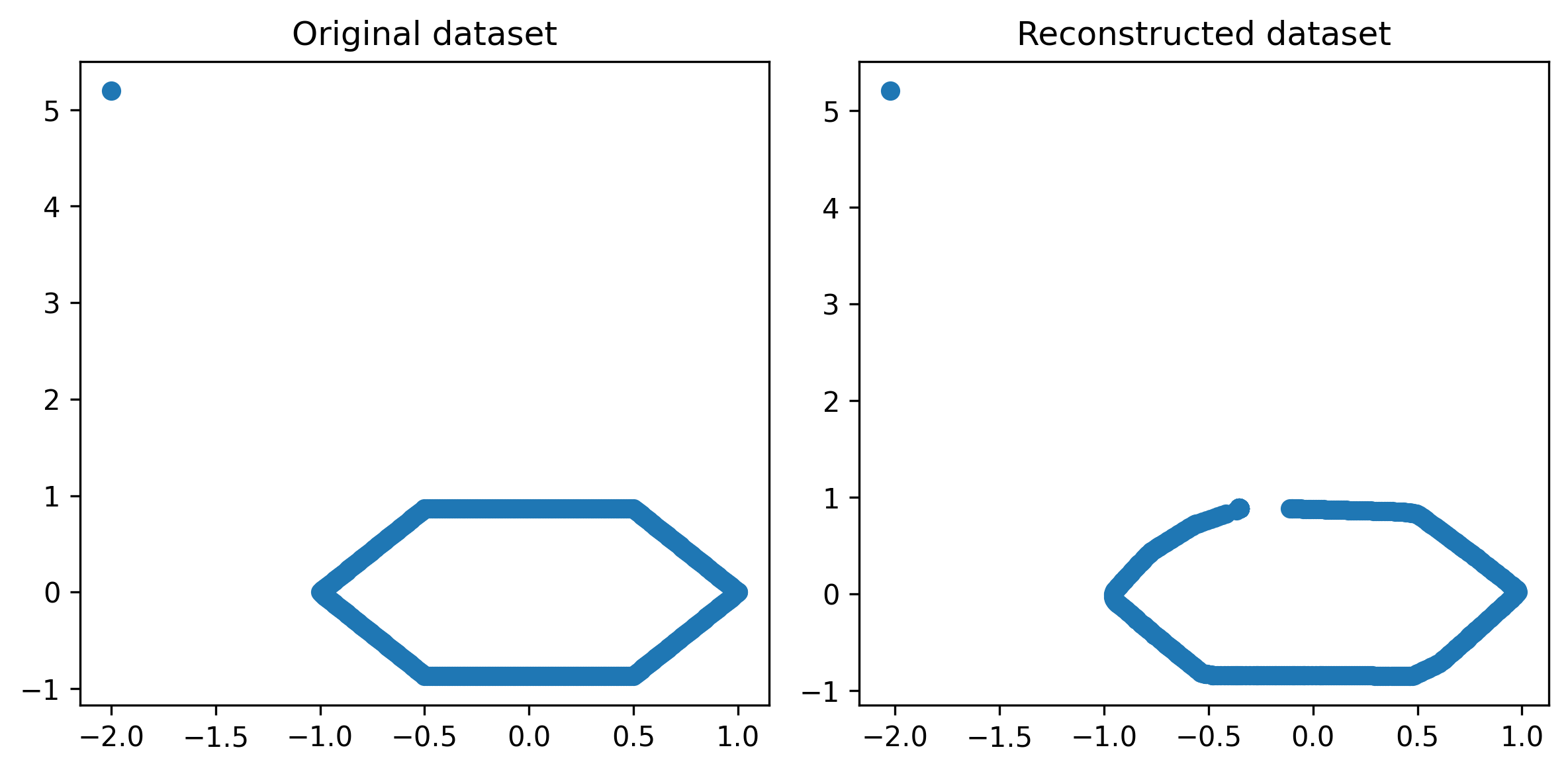}
    \caption{Recreation with learning rate 0.001}
\end{figure}
\begin{figure}[H]
    \includegraphics[width=1\linewidth]{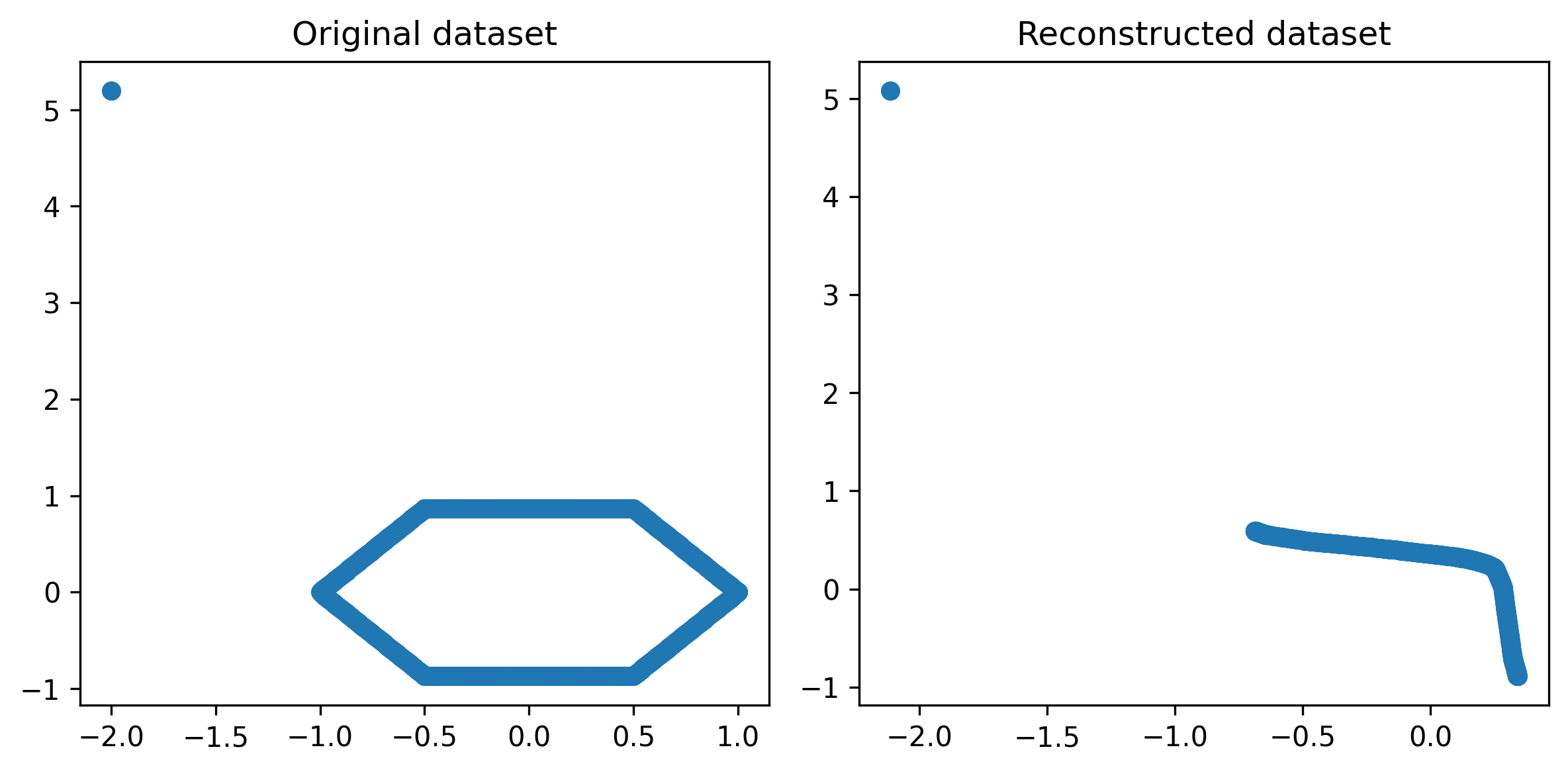}
    \caption{Recreation with learning rate 0.0001}
\end{figure}
\begin{figure}[H]
    \centering
    \includegraphics[width=1\linewidth]{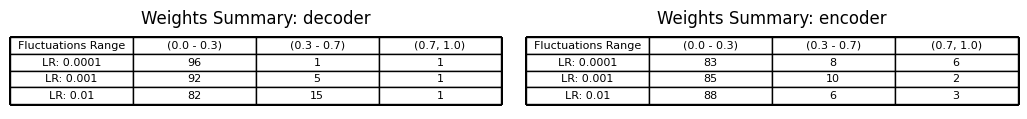}
    \caption{Fluctuations in Weights}
    \label{fig:HeFW}
\end{figure}
\begin{figure}[H]
    \centering
    \includegraphics[width=1\linewidth]{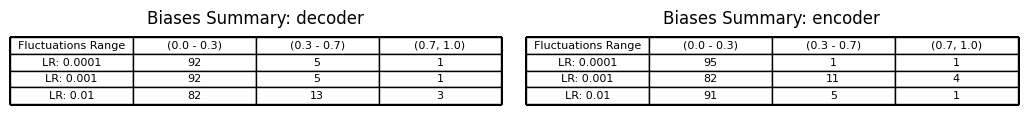}
    \caption{Fluctuations in Biases}
    \label{fig:HeFB}
\end{figure}
\begin{figure}[H]
    \centering
    \includegraphics[width=1\linewidth]{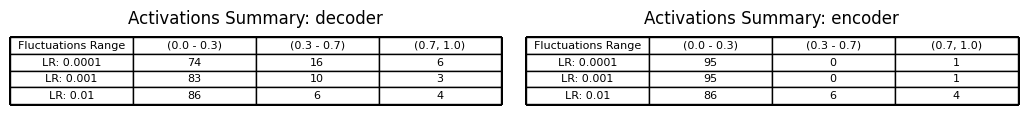}
    \caption{Fluctuations in Activations}
    \label{fig:HeFA}
\end{figure}
\begin{figure}[H]
    \centering
    \includegraphics[width=1\linewidth]{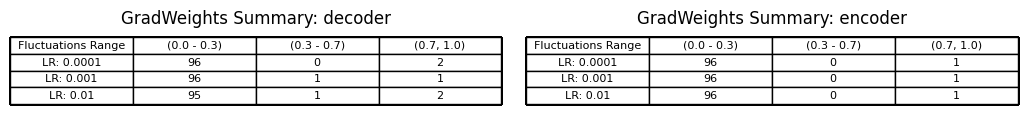}
    \caption{Fluctuations in Weight Gradient}
    \label{fig:HeFGW}
\end{figure}
\begin{figure}[H]
    \centering
    \includegraphics[width=1\linewidth]{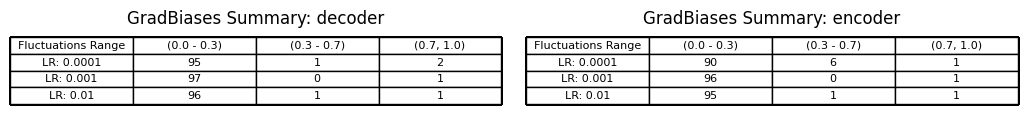}
    \caption{Fluctuations in Bias Gradient}
    \label{fig:HeFGB}
\end{figure}
\subsection{Heptagon}
\begin{figure}[H]
    \centering
    \includegraphics[width=1\linewidth]{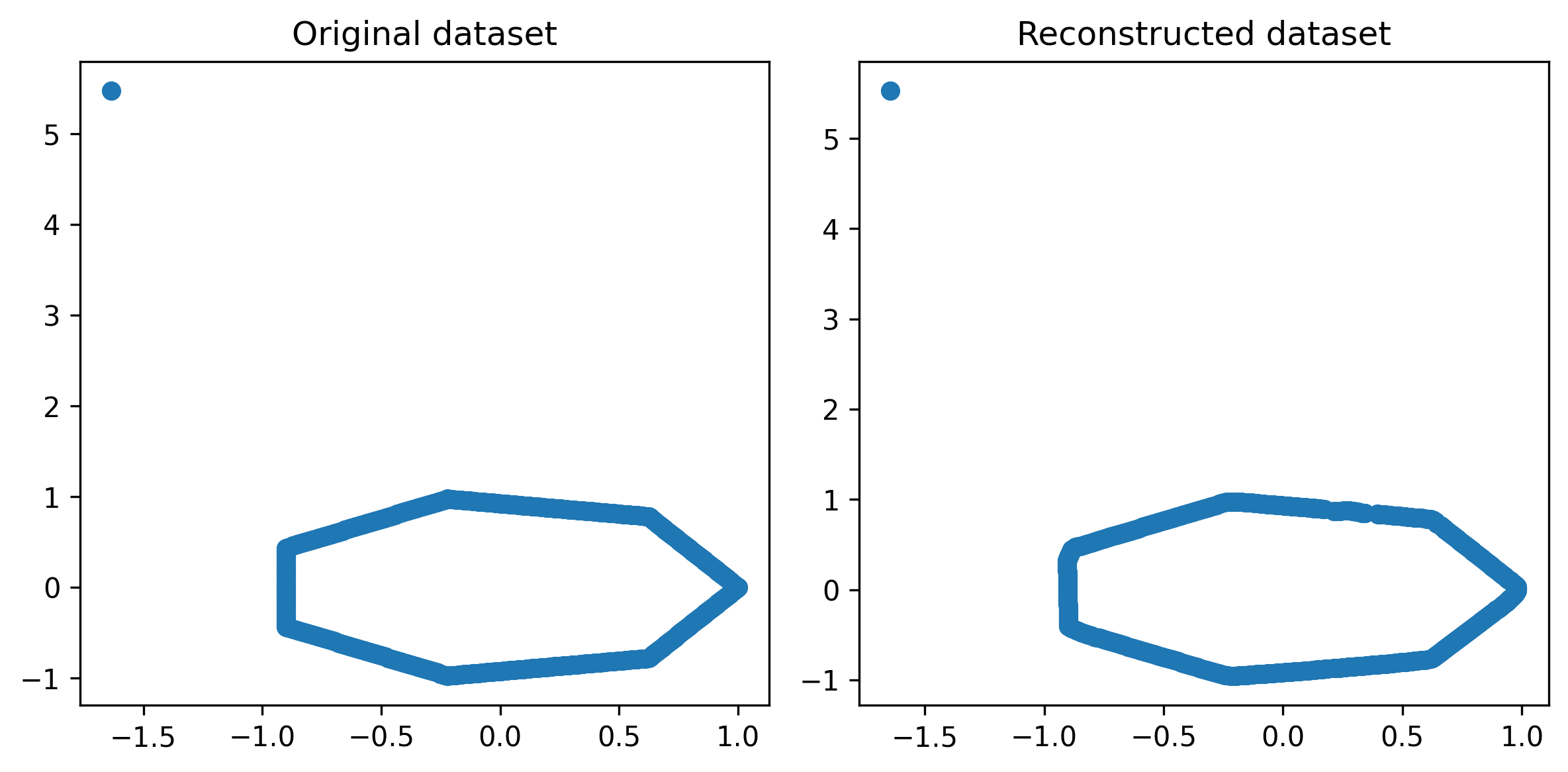}
    \caption{Recreation with learning rate 0.01}
\end{figure}
\begin{figure}[H]
    \includegraphics[width=1\linewidth]{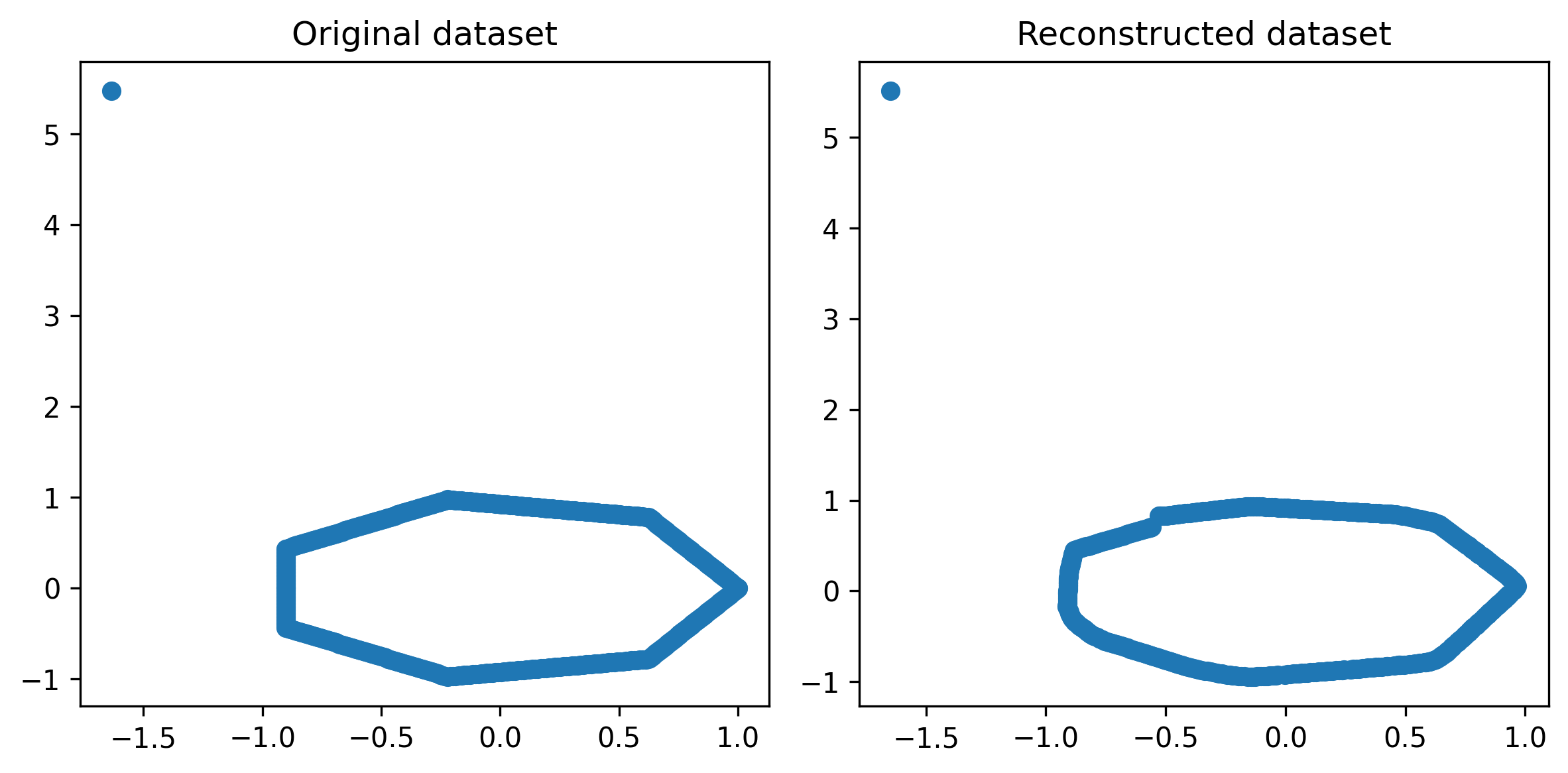}
    \caption{Recreation with learning rate 0.001}
\end{figure}
\begin{figure}[H]
    \includegraphics[width=1\linewidth]{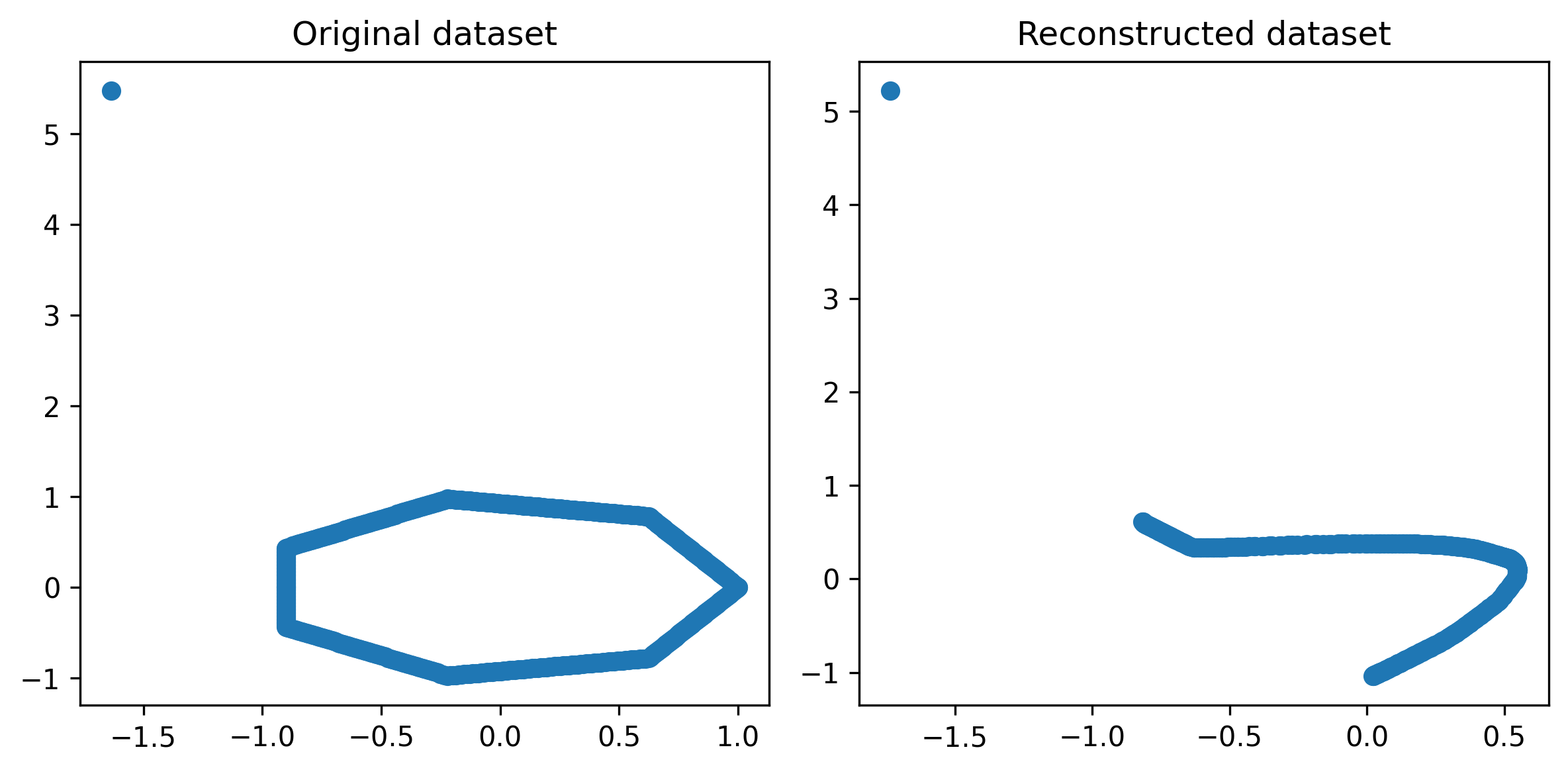}
    \caption{Recreation with learning rate 0.0001}
\end{figure}
\begin{figure}[H]
    \centering
    \includegraphics[width=1\linewidth]{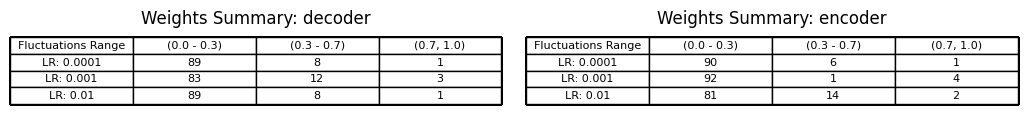}
    \caption{Fluctuations in Weights}
    \label{fig:HFW}
\end{figure}
\begin{figure}[H]
    \centering
    \includegraphics[width=1\linewidth]{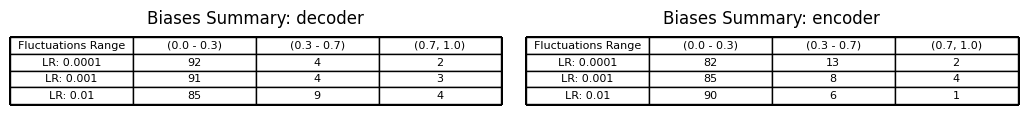}
    \caption{Fluctuations in Biases}
    \label{fig:HFB}
\end{figure}
\begin{figure}[H]
    \centering
    \includegraphics[width=1\linewidth]{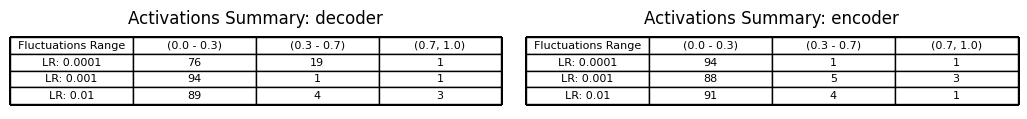}
    \caption{Fluctuations in Activations}
    \label{fig:HFA}
\end{figure}
\begin{figure}[H]
    \centering
    \includegraphics[width=1\linewidth]{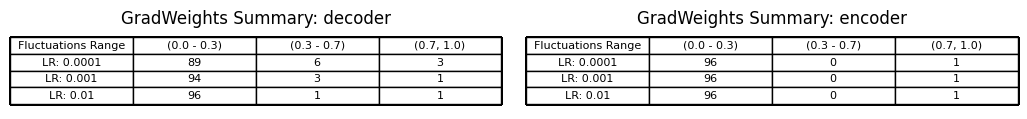}
    \caption{Fluctuations in Weight Gradient}
    \label{fig:HFGW}
\end{figure}
\begin{figure}[H]
    \centering
    \includegraphics[width=1\linewidth]{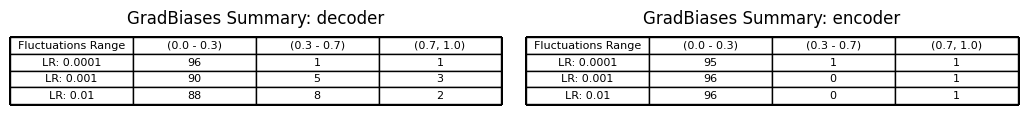}
    \caption{Fluctuations in Bias Gradient}
    \label{fig:HFGB}
\end{figure}
\subsection{Octagon}
\begin{figure}[H]
    \centering
    \includegraphics[width=1\linewidth]{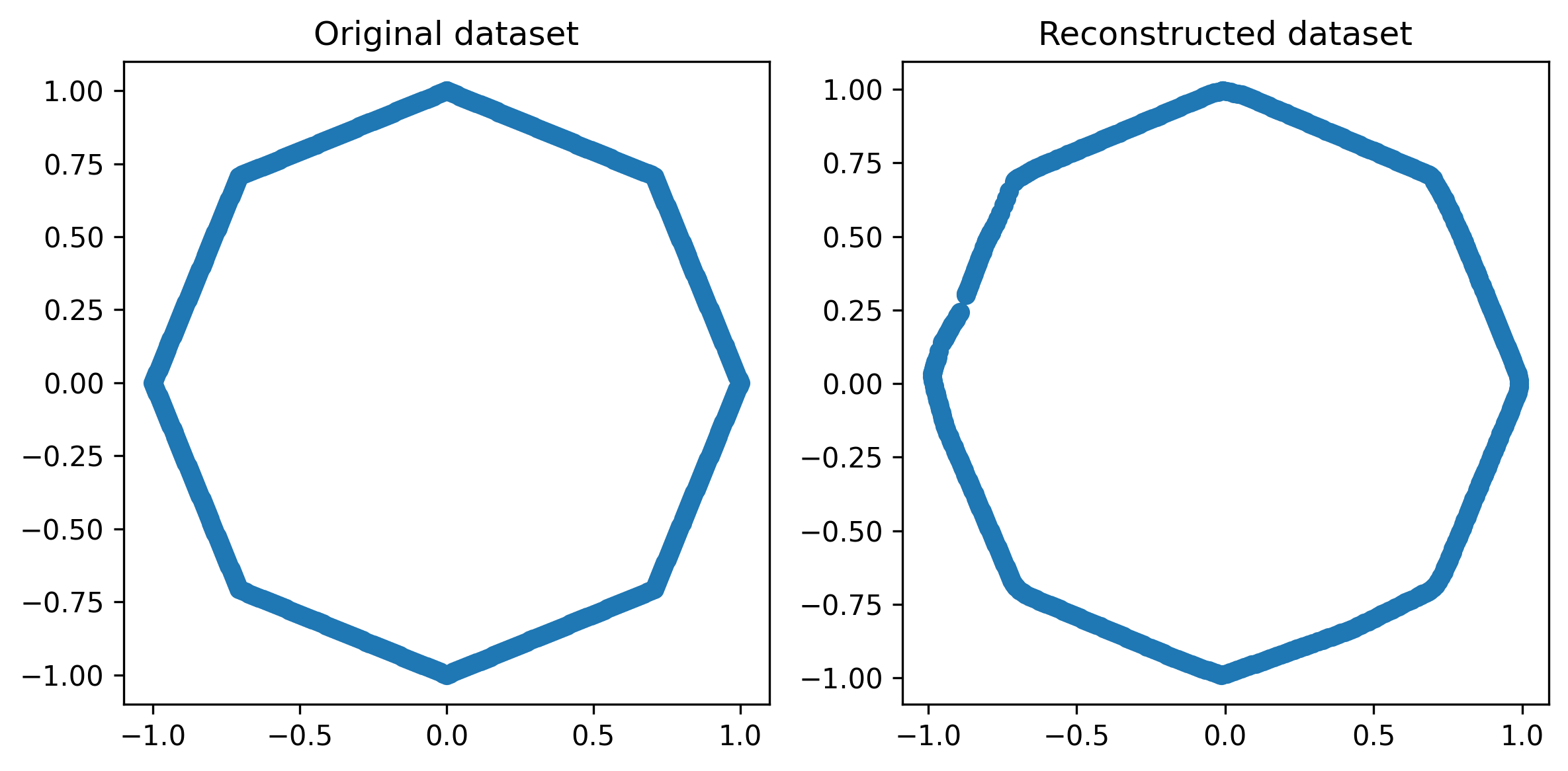}
    \caption{Recreation with learning rate 0.01}
\end{figure}
\begin{figure}[H]
    \includegraphics[width=1\linewidth]{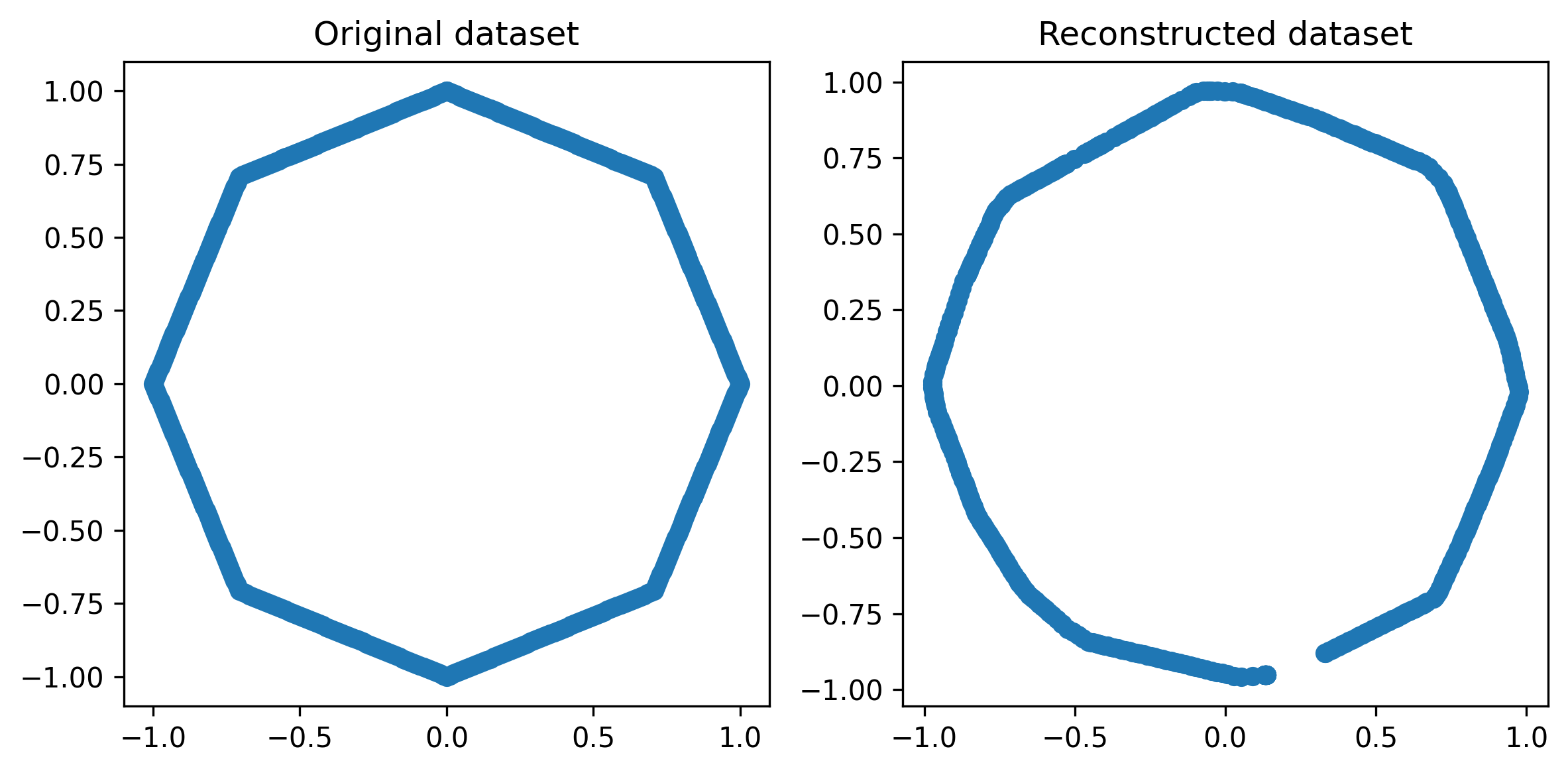}
    \caption{Recreation with learning rate 0.001}
\end{figure}
\begin{figure}[H]
    \includegraphics[width=1\linewidth]{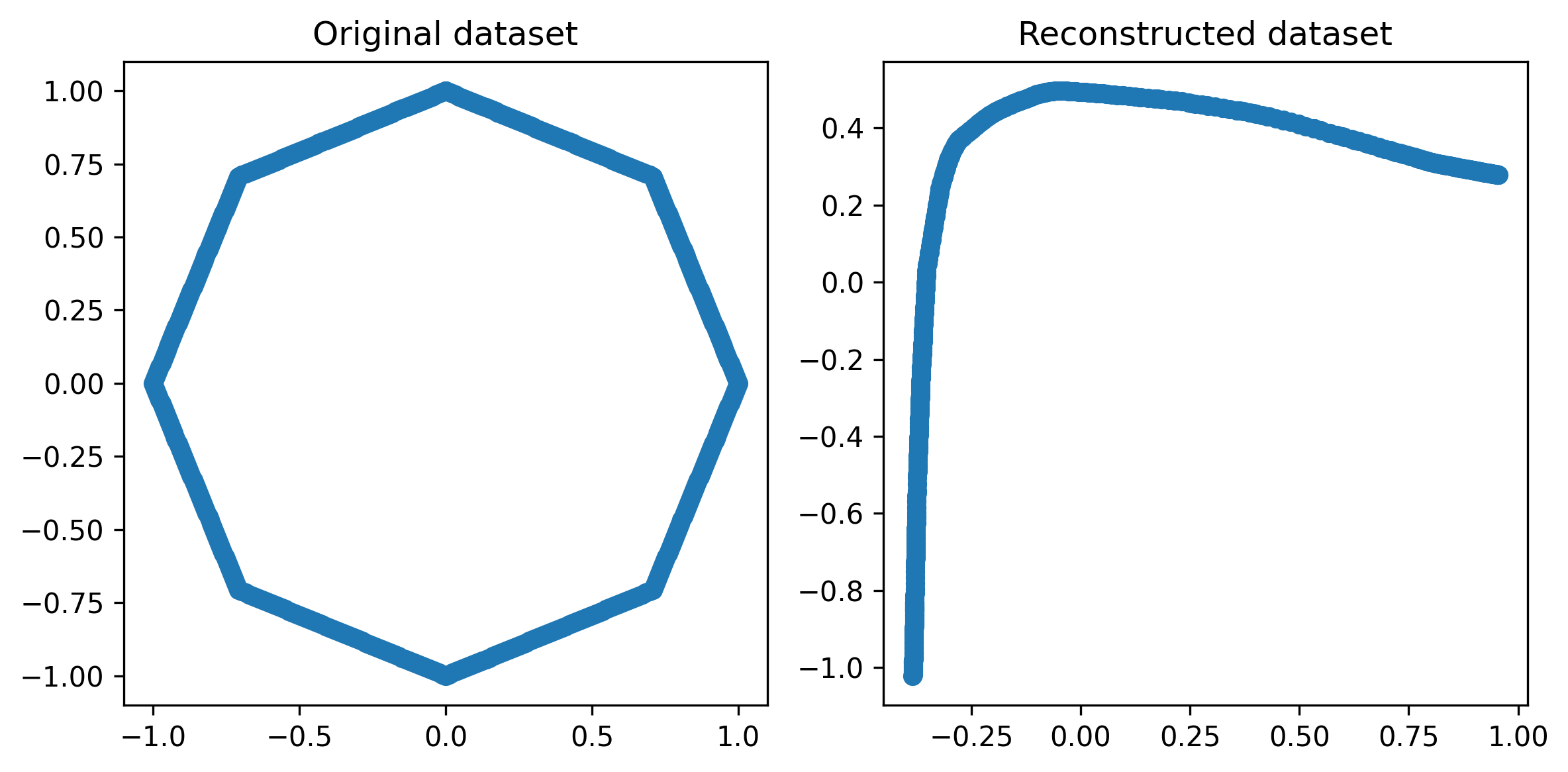}
    \caption{Recreation with learning rate 0.0001}
\end{figure}
\begin{figure}[H]
    \centering
    \includegraphics[width=1\linewidth]{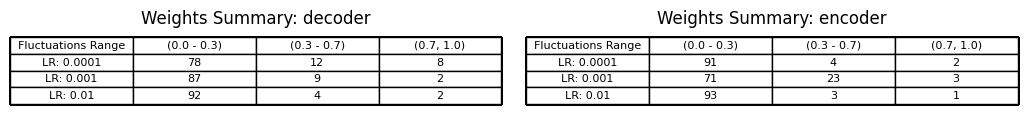}
    \caption{Fluctuations in Weights}
    \label{fig:OFW}
\end{figure}
\begin{figure}[H]
    \centering
    \includegraphics[width=1\linewidth]{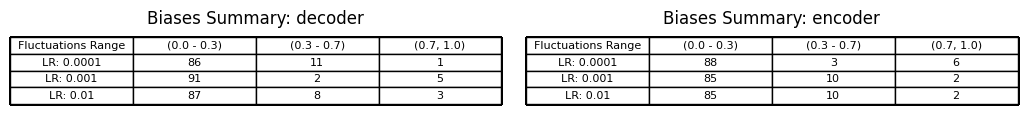}
    \caption{Fluctuations in Biases}
    \label{fig:OFB}
\end{figure}
\begin{figure}[H]
    \centering
    \includegraphics[width=1\linewidth]{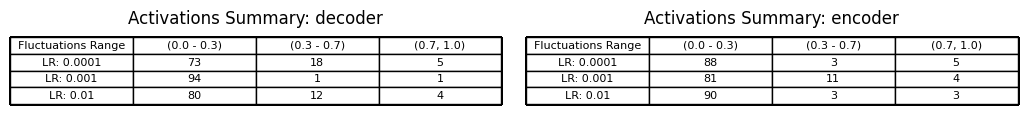}
    \caption{Fluctuations in Activations}
    \label{fig:OFA}
\end{figure}
\begin{figure}[H]
    \centering
    \includegraphics[width=1\linewidth]{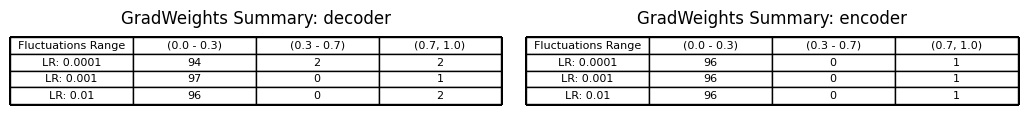}
    \caption{Fluctuations in Weight Gradient}
    \label{fig:OFGW}
\end{figure}
\begin{figure}[H]
    \centering
    \includegraphics[width=1\linewidth]{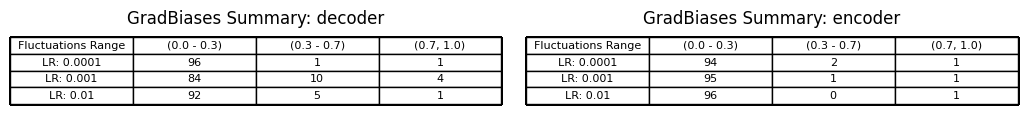}
    \caption{Fluctuations in Bias Gradient}
    \label{fig:OFGB}
\end{figure}
\subsection{Circle}
\begin{figure}[H]
    \centering
    \includegraphics[width=1\linewidth]{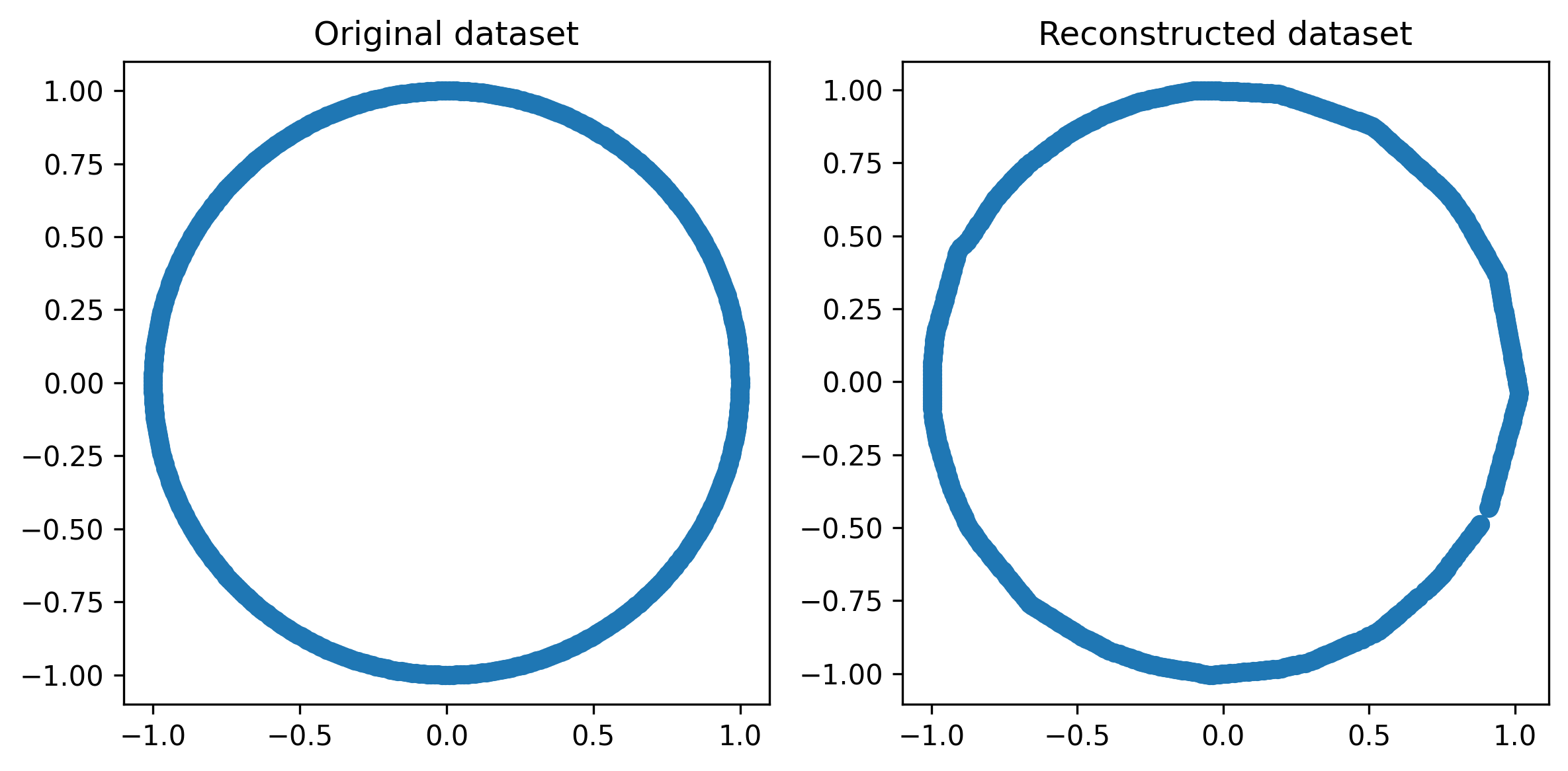}
    \caption{Recreation with learning rate 0.01}
\end{figure}
\begin{figure}[H]
    \includegraphics[width=1\linewidth]{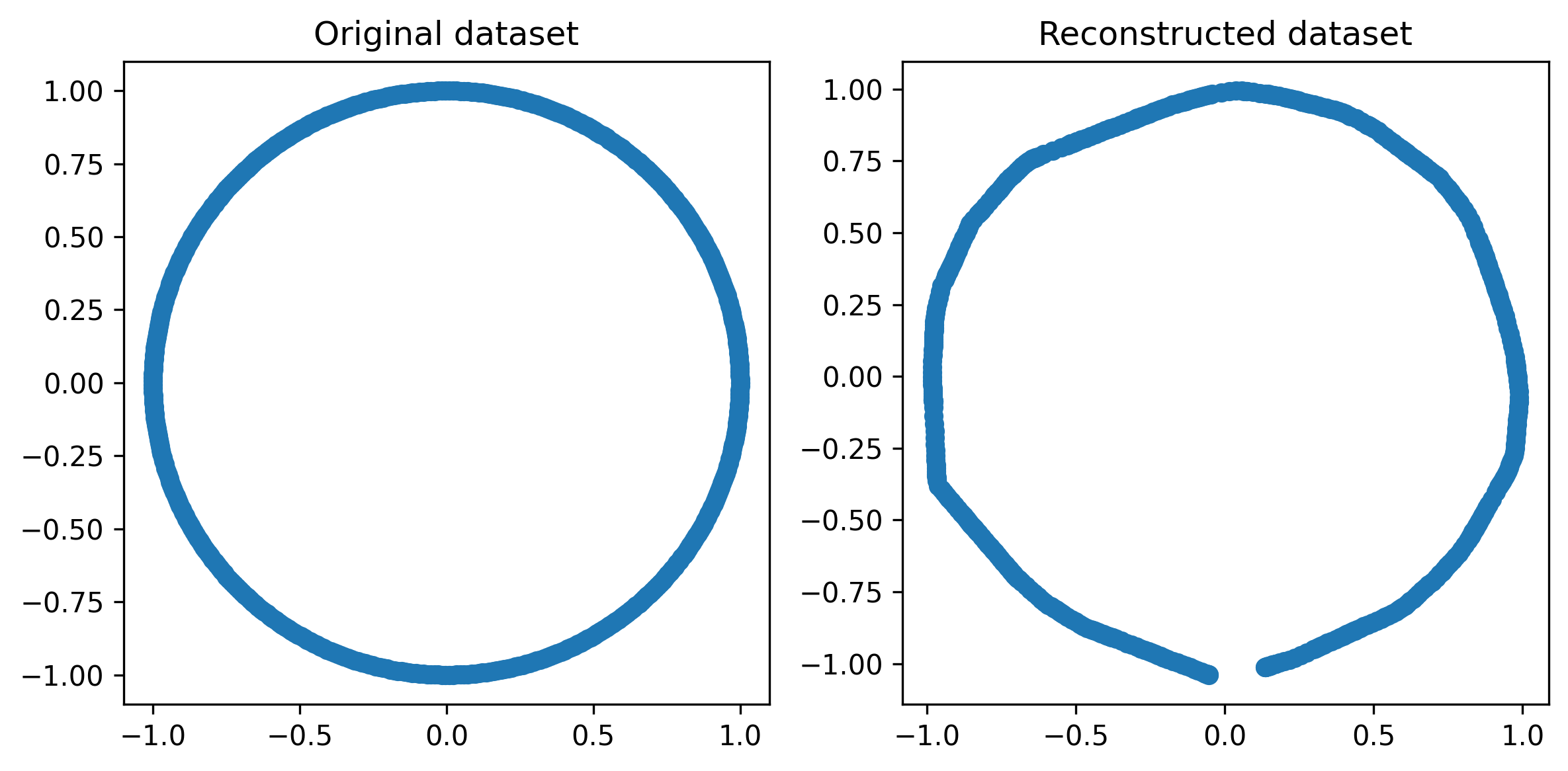}
    \caption{Recreation with learning rate 0.001}
\end{figure}
\begin{figure}[H]
    \includegraphics[width=1\linewidth]{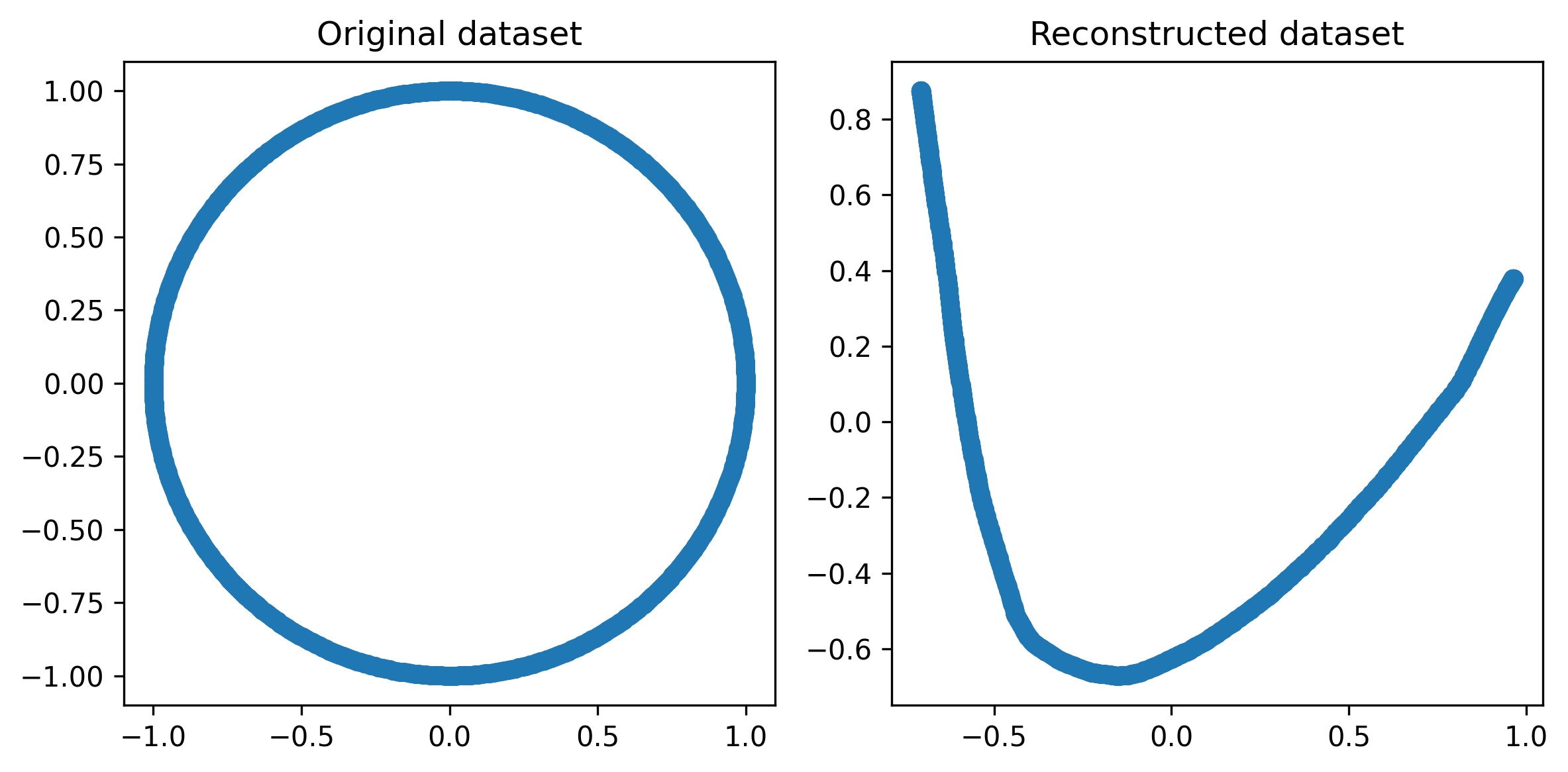}
    \caption{Recreation with learning rate 0.0001}
\end{figure}
\begin{figure}[H]
    \centering
    \includegraphics[width=1\linewidth]{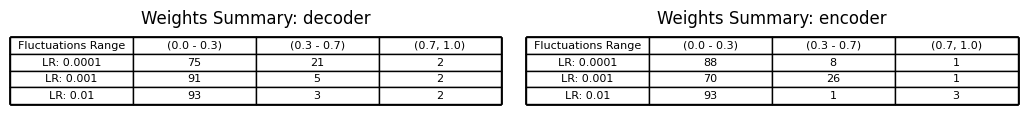}
    \caption{Fluctuations in Weights}
    \label{fig:CFW}
\end{figure}
\begin{figure}[H]
    \centering
    \includegraphics[width=1\linewidth]{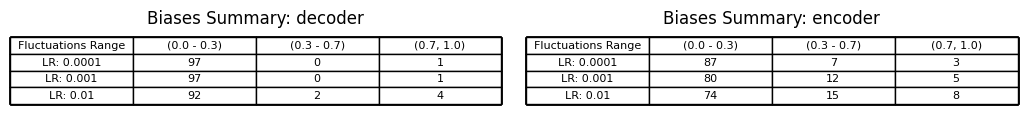}
    \caption{Fluctuations in Biases}
    \label{fig:CFB}
\end{figure}
\begin{figure}[H]
    \centering
    \includegraphics[width=1\linewidth]{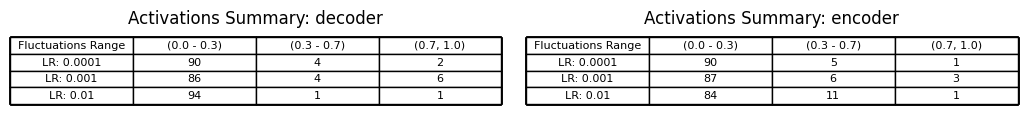}
    \caption{Fluctuations in Activations}
    \label{fig:CFA}
\end{figure}
\begin{figure}[H]
    \centering
    \includegraphics[width=1\linewidth]{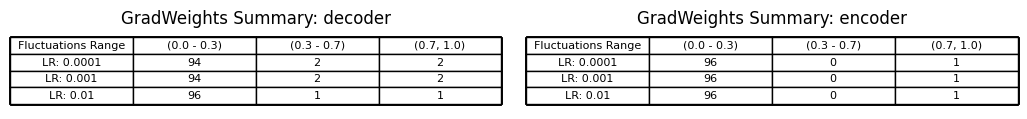}
    \caption{Fluctuations in Weight Gradient}
    \label{fig:CFGW}
\end{figure}
\begin{figure}[H]
    \centering
    \includegraphics[width=1\linewidth]{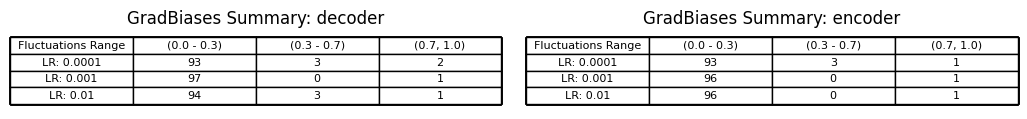}
    \caption{Fluctuations in Bias Gradient}
    \label{fig:CFGB}
\end{figure}
\end{document}